\documentclass[10pt,twocolumn,letterpaper]{article}

\usepackage{cvpr}
\usepackage{times}
\usepackage{epsfig}
\usepackage{graphicx}
\usepackage{tabularx}
\usepackage{amsmath}
\usepackage{amssymb}
\usepackage{color}

\DeclareMathOperator*{\argmax}{arg\,max}
\newcolumntype{Y}{>{\centering\arraybackslash}X}

\definecolor{bananayellow}{rgb}{1.0, 0.75, 0.0}

\usepackage[pagebackref=true,breaklinks=true,letterpaper=true,colorlinks,bookmarks=false]{hyperref}

\cvprfinalcopy 

\def\cvprPaperID{****} 

\ifcvprfinal\fi
\begin{document}

\title{Modeling Relationships in Referential Expressions \\ with Compositional Modular Networks}

\author{Ronghang Hu$^1$ \quad Marcus Rohrbach$^1$ \quad Jacob Andreas$^1$ \quad Trevor Darrell$^1$ \quad Kate Saenko$^2$ \\
$^1$University of California, Berkeley \qquad $^2$Boston University \\
{\tt\small \{ronghang,rohrbach,jda,trevor\}@eecs.berkeley.edu, saenko@bu.edu}}

\maketitle
\thispagestyle{empty}

\begin{abstract}
People often refer to entities in an image in terms of their relationships with other entities. For example, \emph{the black cat sitting under the table} refers to both a \emph{black cat} entity and its relationship with another \emph{table} entity. Understanding these relationships is essential for interpreting and grounding such natural language expressions. Most prior work focuses on either grounding entire referential expressions holistically to one region, or localizing relationships based on a fixed set of categories. In this paper we instead present a modular deep architecture capable of analyzing referential expressions into their component parts, identifying entities and relationships mentioned in the input expression and grounding them all in the scene. We call this approach Compositional Modular Networks (CMNs): a novel architecture that learns linguistic analysis and visual inference end-to-end. Our approach is built around two types of neural modules that inspect local regions and pairwise interactions between regions. We evaluate CMNs on multiple referential expression datasets, outperforming state-of-the-art approaches on all tasks.
\end{abstract}

\section{Introduction}

\begin{figure}[t]
\begin{center}
\includegraphics[width=0.95\linewidth]{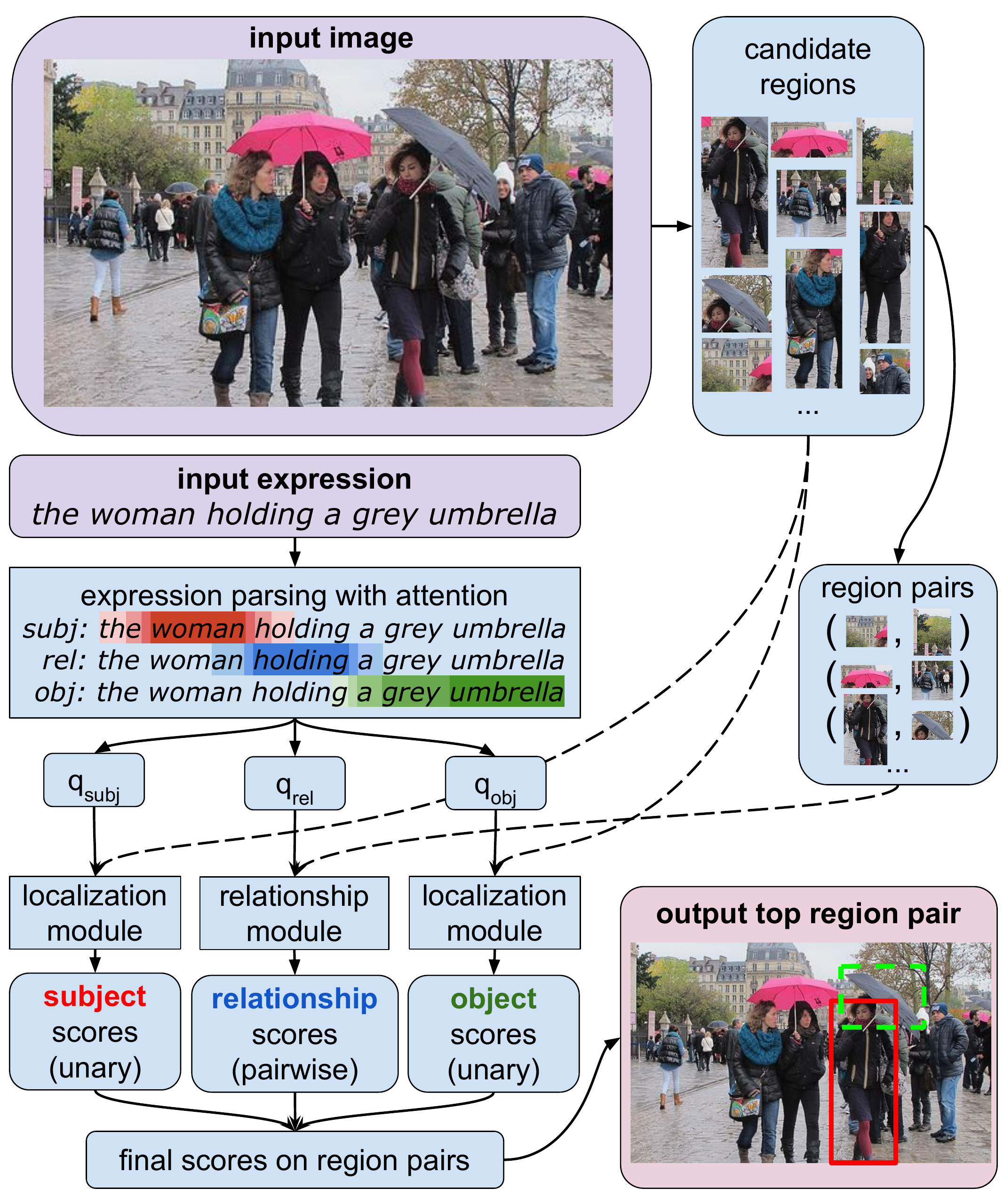}
\end{center}
\caption{Given an image and an expression, we learn to parse the expression into vector representation of subject $q_{subj}$, relationship $q_{rel}$ and object $q_{obj}$ with attention, and align these textual components to image regions with two types of modules. The localization module outputs scores over each individual region while the relationship module produces scores over region pairs. These outputs are integrated into final scores over region pairs, producing the top region pair as grounding result. (Best viewed in color.)
}
\label{fig:teaser}
\end{figure}

Great progress has been made on object detection, the task of localizing visual entities belonging to a pre-defined set of categories \cite{girshick2014rich,ren2015faster,redmon2016you,dai2016r,liu2015ssd}. But the more general and challenging task of localizing entities based on arbitrary natural language expressions remains far from solved.
This task, sometimes known as \emph{grounding} or \emph{referential expression comprehension}, has been explored by recent work in both computer vision and natural language processing \cite{mao2016generation,hu2016natural,rohrbach2016grounding}. Given an image and a natural language expression referring to a visual entity, such as \emph{the young man wearing green shirt and riding a black bicycle}, these approaches localize the image region corresponding to the entity that the expression refers to with a bounding box.

Referential expressions often describe relationships between multiple entities in an image. In Figure \ref{fig:teaser}, for example, the expression \emph{the woman holding a grey umbrella} describes a \emph{woman} entity that participates in a \emph{holding} relationship with a \emph{grey umbrella} entity. Because there are multiple women in the image, resolving this referential expression requires both finding a bounding box that contains a person, and ensuring that this bounding box relates in the right way to other objects in the scene.
Previous work on grounding referential expressions either (1) treats referential expressions holistically, thus failing to model explicit correspondence between textual components and visual entities in the image \cite{mao2016generation,hu2016natural,rohrbach2016grounding,yu2016modeling,nagaraja2016modeling}, or else (2) relies on a fixed set of entity and relationship categories defined \emph{a priori} \cite{lu2016visual}. 

In this paper, we present a joint approach that explicitly models the compositional linguistic structure of referential expressions and their groundings, but which nonetheless supports interpretation of arbitrary language.
We focus on referential expressions involving inter-object relationships that can be represented as a subject entity, a relationship and an object entity. We propose Compositional Modular Networks (CMNs), an end-to-end trained model that learns language representation and image region localization jointly as shown in Figure \ref{fig:teaser}. Our model differentiably parses the referential expression into a subject, relationship and object with three soft attention maps, and aligns the extracted textual representations with image regions using a modular neural architecture. There are two types of modules in our model, one used for localizing specific textual components by outputting unary scores over regions for that component, and one for determining the relationship between two pairs of bounding boxes by outputting pairwise scores over region-region pairs.
We evaluate our model on multiple datasets containing referential expressions, and show that our model outperforms both natural baselines and previous work.

\section{Related work}

\textbf{Grounding referential expressions.} The problem of grounding referential expressions can be naturally formulated as a retrieval problem over image regions \cite{mao2016generation,hu2016natural,rohrbach2016grounding,fukui2016multimodal,yu2016modeling,nagaraja2016modeling}. First, a set of candidate regions are extracted (\eg via object proposal methods like \cite{uijlings2013selective,arbelaez2014multiscale,krahenbuhl2014geodesic,zitnick2014edge}). Next, each candidate region is scored by a model with respect to the query expression, returning the highest scoring candidate as the grounding result. In \cite{mao2016generation,hu2016natural}, each region is scored based on its local visual features and some global contextual features from the whole image. However, local visual features and global contextual from the whole image are often insufficient to determine whether a region matches an expression, as relationships with other regions in the image must also be considered.
Two recent methods \cite{yu2016modeling,nagaraja2016modeling} go beyond local visual features in a single region, and consider multiple regions at the same time. \cite{yu2016modeling} adds contextual feature extracted from other regions in the image, and \cite{nagaraja2016modeling} proposes a model that grounds a referential expression into a pair of regions. All these methods represent language holistically using a recurrent neural network: either generatively, by predicting a distribution over referential expressions \cite{mao2016generation,hu2016natural,yu2016modeling,nagaraja2016modeling}, or discriminatively, by encoding expressions into a vector representation \cite{rohrbach2016grounding,fukui2016multimodal}. This makes it difficult to learn explicit correspondences between the components in the textual expression and entities in the image. In this work, we learn to parse the language expression into textual components in instead of treating it as a whole, and align these components with image regions end-to-end.

\textbf{Handling inter-object relationships.} Recently work by \cite{lu2016visual} trains detectors based on RCNN \cite{girshick2014rich} and uses a linguistic prior to detect visual relationships. However, this work relies on fixed, predefined categories for subjects, relations, and objects, treating entities like ``bicycle'' and relationships like and ``riding'' as discrete classes. Instead of building upon a fixed inventory of classes, our model handles relationships specified by arbitrary natural language phrases, and jointly learns expression parsing and visual entity localization. Although \cite{krishnamurthy2013jointly} also learns language parsing and perception, it is directly based on logic ($\lambda$-calculus) and requires additional classifiers trained for each predicate class.

\textbf{Compositional structure with modules.} Neural Module Networks \cite{andreas16cvpr} address visual question answering by decomposing the questions into textual components and dynamically assembling a specific network architecture for the question from a few network modules based on the textual components. However, this method relies on an external language parser for textual analysis instead of end-to-end learned language representation, and is not directly applicable to the task of grounding referential expressions into bounding boxes, since it does not explicitly output bounding boxes as results. Recently, \cite{andreas2016learning} improves over \cite{andreas16cvpr} by learning to re-rank parsing outputs from the external parser, but it is still not end-to-end learned since the parser is fixed and not optimized for the task. Inspired by \cite{andreas16cvpr}, our model also uses a modular structure, but learns the language representation end-to-end from words.

\section{Our model}

\begin{figure*}[t]
\begin{center}
\includegraphics[width=0.95\linewidth]{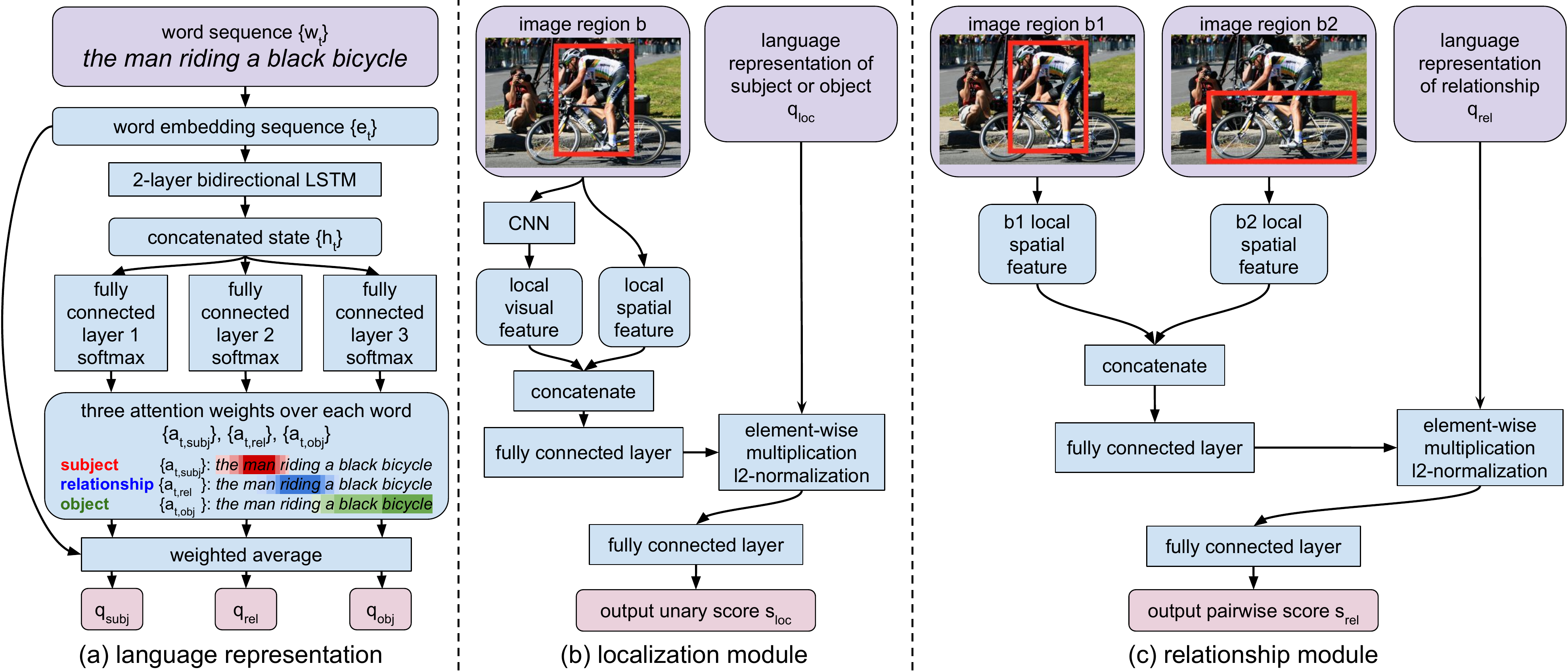}
\end{center}
\caption{Detailed illustration of our model. (a) Our model learns to parse an expression into subject, relationship and object with attention for language representation (Sec. \ref{sec:lang_representation}). (b) The localization module matches subject or object with each image region and returns a unary score (Sec. \ref{sec:loc_module}). (c) The relationship module matches a relationship with a pair of regions and returns a pairwise score (Sec. \ref{sec:rel_module}).}
\label{fig:method}
\end{figure*}

We propose Compositional Modular Networks (CMNs) to localize visual entities described by a query referential expression. Our model is compositional in the sense that it localizes a referential expression by grounding the components in the expressions and exploiting their interactions, in accordance with the principle of compositionality of natural language -- the meaning of a complex expression is determined by the meanings of its constituent expressions and the rules used to combine them \cite{werning2012oxford}.
Our model works in a retrieval setting: given an image $I$, a referential expression $Q$ as query and a set of candidate region bounding boxes $B=\{b_i\}$ for the image $I$ (\eg extracted through object proposal methods), our model outputs a score for each bounding box $b_i$, and returns the bounding box with the highest score as grounding (localization) result. Unlike state-of-the-art methods \cite{rohrbach2016grounding,fukui2016multimodal}, the scores for each region bounding box $b_i \in B$ are not predicted only from the local feature of $b_i$, but also based on other regions in the image. In our model, we focus on the relationships in referential expressions that can be represented as a 3-component triplet \texttt{(subject, relationship, object)}, and learn to parse the expressions into these components with attention. For example, \emph{a young man wearing a blue shirt} can be parsed as the triplet (\emph{a young man}, \emph{wearing}, \emph{a blue shirt}). The score of a region is determined by simultaneously looking at whether it matches the description of the subject entity and whether it matches the relationship with another interacting object entity mentioned in the expression.

Our model handles such inter-object relationships by looking at pairs of regions $(b_i, b_j)$. For referential expressions like ``the red apple on top of the bookshelf'', we want to find a region pair $(b_i, b_j)$ such that $b_i$ matches the subject entity ``red apple'' and $b_j$ matches the object entity ``bookshelf'' and the configuration of $(b_i, b_j)$ matches the relationship ``on top of''. To achieve this goal, our model is based on a compositional modular structure, composed of two modules assembled in a pipeline for different sub-tasks: one localization module $f_{loc}(\cdot, q_{loc}; \Theta_{loc})$ for deciding whether a region matches the subject or object in the expression, where $q_{loc}$ is the textual vector representation of the subject component ``red apple'' or the object component ``bookshelf'', and one relationship module $f_{rel}(\cdot, \cdot, q_{rel}; \Theta_{rel})$ for deciding whether a pair of regions matches the relationship described in the expression represented by $q_{rel}$, the textual vector representation of the relationship ``on top of''. The representations $q_{subj}$, $q_{rel}$ and $q_{obj}$ are learned jointly in our model in Sec. \ref{sec:lang_representation}.

We define the pairwise score $s_{pair}(b_i, b_j)$ over a pair of image regions $(b_i, b_j)$ matching an input referential expression $Q$ as the sum of three components:
\begin{equation}\label{eqn:score_pair}
\begin{split}
s_{pair}(b_i, b_j) &= f_{loc}(b_i, q_{subj}; \Theta_{loc}) \\
&+ f_{loc}(b_j, q_{obj}; \Theta_{loc}) \\
&+ f_{rel}(b_i, b_j, q_{rel}; \Theta_{rel}),
\end{split}
\end{equation}
where $q_{subj}$, $q_{obj}$ and $q_{rel}$ are vector representations of subject, relationship and object, respectively.

For inference, we define the final subject unary score $s_{subj}(b_i)$ of a bounding of $b_i$ corresponding to the subject (\eg ``the red apple'' in ``the red apple on top of the bookshelf'') as the score of the best possible pair $(b_i, b_j)$ that matches the entire expression:
\begin{equation}\label{eqn:score_unary}
s_{subj}(b_i) \triangleq \max_{b_j \in B}\; s_{pair}(b_i, b_j).
\end{equation}
The subject is ultimately grounded (localized) to the highest scoring region as
\begin{equation}
b^*_{subj} = \argmax_{b_i \in B} (s_{subj}(b_i)).
\end{equation}

\subsection{Expression parsing with attention}\label{sec:lang_representation}

Given a referential expression $Q$ like \emph{the tall woman carrying a red bag}, how can we decide which substrings corresponds to the subject, the relationship, and the object, and extract three vector representations $q_{subj}$, $q_{rel}$ and $q_{obj}$ corresponding to these three components? One possible approach is to use an external language parser to parse the referential expression into the triplet format \texttt{(subject, relationship, object)} and then process each component with an encoder (\eg a recurrent neural network) to extract $q_{subj}$, $q_{rel}$ and $q_{obj}$. 
However, the formal representations of language produced by syntactic parsers
do not always correspond to intuitive visual representations. As a simple
example, \emph{the apple on top of the bookshelf} is analyzed \cite{zhu2013fast} as having a subject
phrase \emph{the apple}, a relationship \emph{on}, and an object phrase \emph{top of the bookshelf}, when in fact the visually salient objects are simply the apple and the bookshelf, while the complete expression \emph{on top of} describes the relationship between them.

Therefore, in this work we learn to decompose the input expression $Q$ into the above 3 components, and generate vector representations $q_{subj}$, $q_{rel}$ and $q_{obj}$ from $Q$ through a soft attention mechanism over the word sequence, as shown in Figure \ref{fig:method} (a). For a referential expression $Q$ that is a sequence of $T$ words $\{w_t\}_{t=1}^T$, we first embed each word $w_t$ to a vector $e_t$ using GloVe \cite{pennington2014glove}, and then scan through the word embedding sequence $\{e_t\}_{t=1}^T$ with a 2-layer bi-directional LSTM network \cite{schuster1997bidirectional}. The first layer takes as input the sequence $\{e_t\}$ and outputs a forward hidden state $h^{(1, fw)}_t$ and a backward hidden state $h^{(1, bw)}_t$ at each time step, which are concatenated into $h^{(1)}_t$. The second layer then takes the first layer's output sequence $\{h^{(1)}_t\}$ as input and outputs forward and backward hidden states $h^{(2, fw)}_t$ and $h^{(2, bw)}_t$ at each time step. All the hidden states in the first layer and second layer are concatenated into a single vector $h_t$.
\begin{equation}
h_t = \left[h^{(1, fw)}_t~~h^{(1, bw)}_t~~h^{(2, fw)}_t~~h^{(2, bw)}_t\right]
\end{equation}

The concatenated state $h_t$ contains information from word $w_t$ itself and also context from words before and after $w_t$. Then the attention weights $a_{t, subj}$, $a_{t, rel}$ and $a_{t, obj}$ for subject, relationship, object over each word $w_t$ are obtained by three linear predictions over $h_t$ followed by a softmax as
\begin{eqnarray}
a_{t, subj} &=& \frac{\exp\left(\beta_{subj}^T h_t\right)}{\sum_{\tau=1}^T \exp\left(\beta_{subj}^T h_\tau\right)}\label{eqn:attention_1} \\
a_{t, rel} &=& \frac{\exp\left(\beta_{rel}^T h_t\right)}{\sum_{\tau=1}^T \exp\left(\beta_{rel}^T h_\tau\right)} \\
a_{t, obj} &=& \frac{\exp\left(\beta_{obj}^T h_t\right)}{\sum_{\tau=1}^T \exp\left(\beta_{obj}^T h_\tau\right)}\label{eqn:attention_3}
\end{eqnarray}
and the language representations of the subject $q_{subj}$, relationship $q_{rel}$ and object $q_{obj}$ are extracted as weighed average of word embedding vectors $\{e_t\}$ with attention weights, as follows:
\begin{eqnarray}
q_{subj} &=& \sum_{t=1}^T a_{t, subj} e_t \label{eqn:q_subj} \\
q_{rel} &=& \sum_{t=1}^T a_{t, rel} e_t \label{eqn:q_rel} \\
q_{obj} &=& \sum_{t=1}^T a_{t, obj} e_t \label{eqn:q_obj}.
\end{eqnarray}

In our implementation, both the forward and the backward LSTM in each layer of the bi-directional LSTM network have 1000-dimensional hidden states, so the final $h_t$ is 4000-dimensional. During training, dropout is added on top of $h_t$ as regularization.

\subsection{Localization module}\label{sec:loc_module}

As shown in Figure \ref{fig:method} (b), the localization module $f_{loc}$ outputs a score $s_{loc} = f_{loc}(b, q_{loc}; \Theta_{loc})$ representing how likely a region bounding box $b$ matches $q_{loc}$, which is either the subject textual vector $q_{subj}$ in Eqn. \ref{eqn:q_subj} or object textual vector $q_{obj}$ in Eqn. \ref{eqn:q_obj}.

This module takes the local visual feature $x_{vis}$ and spatial feature $x_{spatial}$ of image region $b$. We extract visual feature $x_{v}$ from image region $b$ using a convolutional neural network \cite{simonyan2015very}, and extract a 5-dimensional spatial feature $x_{s}=[\frac{x_{min}}{W_I}, \frac{y_{min}}{H_I}, \frac{x_{max}}{W_I}, \frac{y_{max}}{H_I}, \frac{S_b}{S_I}]$ from $b$ using the same representation as in \cite{mao2016generation}, where $[x_{min}, y_{min}, x_{max}, y_{max}]$ and $S_b$ are bounding box coordinates and area of $b$, and $W_I$, $H_I$ and $S_I$ are width, height and area of the image $I$. Then, $x_{v}$ and $x_{s}$ are concatenated into a vector $x_{v,s} = [x_{v} ~x_{s}]$ as representation of region $b$.

Since element-wise multiplication is shown to be a powerful way to combine representations from different modalities \cite{ba2014multiple}, we adopt it here to obtain a joint vision and language representation. In our implementation, $x_{v,s}$ is first embedded to a new vector $\tilde{x}_{v,s}$ that has the same dimension as $q_{loc}$ (which is either $q_{subj}$ in Eqn. \ref{eqn:q_subj} or $q_{obj}$ in Eqn. \ref{eqn:q_obj}) through a linear transform, and then element-wise multiplied with $q_{loc}$ to obtain a vector $z_{loc}$, which is L2-normalized into $\hat{z}_{loc}$ to obtain a more robust representation, as follows:
\begin{eqnarray}
\tilde{x}_{v,s} &=& W_{v,s} x_{v,s} + b_{v,s} \\
z_{loc} &=& \tilde{x}_{v,s} \odot q_{loc} \\
\hat{z}_{loc} &=& z_{loc} / \|z_{loc}\|_2
\end{eqnarray}
where $\odot$ is element-wise multiplication between two vectors. Then the score $s_{loc}$ is predicted linearly from $\hat{z}_{loc}$ as
\begin{equation}\label{eqn:s_loc}
s_{loc} = w_{loc}^T \hat{z}_{loc} + b_{loc}.
\end{equation}
The parameters in $\Theta_{loc}$ are $(W_{v,s}, b_{v,s}, w_{loc}, b_{loc})$.

\subsection{Relationship module}\label{sec:rel_module}

As shown in Figure \ref{fig:method} (c), the relationship module $f_{rel}$ outputs a score $s_{rel} = f_{rel}(b_1, b_2, q_{rel}; \Theta_{rel})$ representing how likely a pair of region bounding boxes $(b_1, b_2)$ matches $q_{rel}$, the representation of relationship in the expression.

In our implementation, we use the spatial features $x_{s1}$ and $x_{s2}$ of the two regions $b_1$ and $b_2$ extracted in the same way as in localization module (we empirically find that adding visual features of $b_1$ and $b_2$ leads to no noticeable performance boost while slowing training significantly). Then $x_{s1}$ and $x_{s2}$ are concatenated as $x_{s1,s2} = [x_{s1} ~x_{s2}]$, and then processed in a similar way as in localization module to obtain $s_{rel}$, as shown below:
\begin{eqnarray}
\tilde{x}_{s1,s2} &=& W_{s1,s2} x_{s1,s2} + b_{s1,s2} \\
z_{rel} &=& \tilde{x}_{s1,s2} \odot q_{rel} \\
\hat{z}_{rel} &=& z_{rel} / \|z_{rel}\|_2 \\
s_{rel} &=& w_{rel}^T \hat{z}_{rel} + b_{rel}.
\end{eqnarray}
The parameters in $\Theta_{rel}$ are $(W_{s1,s2}, b_{s1,s2}, w_{rel}, b_{rel})$.

\subsection{End-to-end learning}\label{sec:learning}

During training, for an image $I$, a referential expression $Q$ and a set of candidate regions $B$ extracted from $I$, if the ground-truth regions $b_{subj\_gt}$ of the subject entity and $b_{obj\_gt}$ of the object entity are both available, then we can optimize the pairwise score $s_{pair}$ in Eqn. \ref{eqn:score_pair} with strong supervision using softmax loss $Loss_{strong}$.
\begin{equation}\label{eqn:strong_supervision}
Loss_{strong} = -\log\left(\frac{\exp\left(s_{pair}(b_{subj\_gt}, b_{obj\_gt})\right)}{\sum_{(b_i, b_j) \in B \times B} \exp\left(s_{pair}(b_i, b_j)\right)}\right)
\end{equation}
However, it is often hard to obtain ground-truth regions for both subject entity and object entity. For referential expressions like ``a red vase on top of the table'', often there is only a ground-truth bounding box annotation $b_1$ for the subject (vase) in the expression, but no bounding box annotation $b_2$ for the object (table), so one cannot directly optimize the pairwise score $s_{pair}(b_1, b_2)$. To address this issue, we treat the object region $b_2$ as a latent variable, and optimize the unary score $s_{subj}(b_1)$ in Eqn. \ref{eqn:score_unary}. Since $s_{subj}(b_1)$ is obtained by maximizing over all possible region $b_2 \in B$ in $s_{pair}(b_1, b_2)$, this can be regarded as a weakly supervised Multiple Instance Learning (MIL) approach similar to \cite{nagaraja2016modeling}. The unary score $s_{subj}$ can be optimized with weak supervision using softmax loss $Loss_{weak}$.
\begin{equation}\label{eqn:weak_supervision}
Loss_{weak} = -\log\left(\frac{\exp\left(s_{subj}(b_{subj\_gt})\right)}{\sum_{b_i \in B} \exp\left(s_{subj}(b_i)\right)}\right)
\end{equation}

The whole system is trained end-to-end with backpropagation. In our experiments, we train for 300000 iterations, with 0.95 momentum and an initial learning rate of 0.005, multiplied by 0.1 after every 120000 iterations. Each batch contains one image with all referential expressions annotated over that image. Parameters in the localization module, the relationship module and the language representation in our model are initialized randomly with Xavier initializer \cite{glorot2010understanding}. Our model is implemented using TensorFlow \cite{tensorflow2015-whitepaper} and we plan to release our code and data to facilitate reproduction of our results.

\section{Experiments}

We first evaluate our model on a synthetic dataset to verify its ability to handle inter-object relationships in referential expressions. Next we apply our method to real images and expressions in the Visual Genome dataset \cite{krishna2016visual} and Google-Ref dataset \cite{mao2016generation}. Since the task of answering pointing questions in visual question answering is similar to grounding referential expressions, we also evaluate our model on the pointing questions in the Visual-7W dataset \cite{zhu2015visual7w}.

\subsection{Analysis on a synthetic dataset}\label{sec:exp_shape}

\begin{table}[t]
\begin{center}
\begin{tabular}{|l|c|}
\hline
Method & Accuracy \\
\hline
baseline (loc module) & 46.27\% \\
our full model & 99.99\% \\
\hline
\end{tabular}
\end{center}
\caption{Accuracy of our model and the baseline on the synthetic shape dataset. See Sec. \ref{sec:exp_shape} for details.}
\label{tab:results_shape}
\end{table}

\begin{figure}[t]
\begin{center}
\footnotesize{expression=\textit{``the green square right of a red circle''}} \\
\begin{tabularx}{\linewidth}{YYYY}
 & baseline - $s_{loc}$ & $s_{subj}$ & $s_{obj}$ \\
\includegraphics[width=\linewidth]{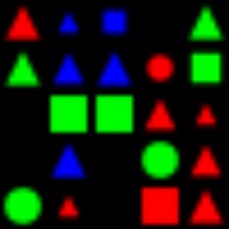} &
\includegraphics[width=\linewidth]{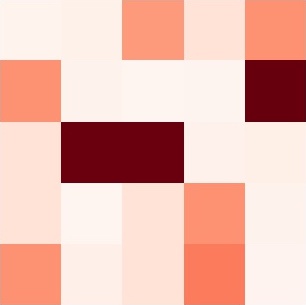} &
\includegraphics[width=\linewidth]{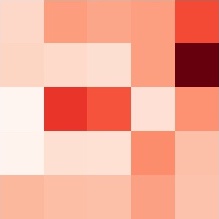} &
\includegraphics[width=\linewidth]{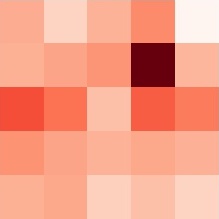} \\
(a) & (b) & (c) & (d) \\
\end{tabularx}
\end{center}
\caption{For the image in (a) and the expression ``the green square right of a red circle'', (b) baseline scores on each location on the 5 by 5 grid using localization module only (darker is higher), and (c, d) scores $s_{subj}$ and $s_{obj}$ using our full model. $s_{subj}$ is highest on the exact green square that is on the right of a red circle, and $s_{obj}$ is highest on this red circle.}
\label{fig:vis_shape}
\end{figure}

Inspired by \cite{andreas16cvpr}, we first perform a simulation experiment on a synthetic shape dataset. The dataset consists of 30000 images with simple circles, squares and triangles of different sizes and colors on a 5 by 5 grid, and referential expressions constructed using a template of the form \texttt{[subj]} \texttt{[relationship]} \texttt{[obj]}, where \texttt{[subj]} and \texttt{[obj]} involve both shape classes and attributes and \texttt{[relationship]} is some spatial relationships such as ``above''. The task is to localize the corresponding shape region described by the expression on the 5 by 5 grid. Figure \ref{fig:vis_shape} (a) shows an example in this dataset with the synthetic expression ``the green square right of a red circle''. In the synthesizing procedure, we make sure that the shape region being referred to cannot be inferred simply from \texttt{[subj]} as there will be multiple matching regions, and the relationship with another region described by \texttt{[obj]} has to be taken into consideration.

On this dataset, we train our model with weak supervision by Eqn. \ref{eqn:weak_supervision} using the ground-truth subject region $b_{subj\_gt}$ of the subject shape described in the expression. Here the candidate region set $B$ are the 25 possible locations on the 5 by 5 grid, and visual features are extracted from the corresponding cropped image region with a VGG-16 network \cite{simonyan2015very} pretrained on ImageNET classification. As a comparison, we also train a baseline model using only the localization module, with a softmax loss on its output $s_{loc}$ in Eqn. \ref{eqn:s_loc} over all 25 locations on the grid, and language representation $q_{loc}$ obtained by scanning through the word embedding sequence with a single LSTM network and taking the hidden state at the last time step same as in \cite{rohrbach2016grounding,hu2016segmentation}. This baseline method resembles the supervised version of GroundeR \cite{rohrbach2016grounding}, and the main difference between this baseline and our model is that the baseline only looks at a region's appearance and spatial property but ignores pairwise relationship with other regions.

We evaluate with the accuracy on whether the predicted subject region $b^*_{subj}$ matches the ground-truth region $b_{subj\_gt}$. Table \ref{tab:results_shape} shows the results on this dataset, where our model trained with weak supervision (the same as the supervision given to baseline) achieves nearly perfect accuracy---significantly outperforming the baseline using a localization module only. Figure \ref{fig:vis_shape} shows an example, where the baseline can localize green squares but fails to distinguish the exact green square right of a red circle, while our model successfully finds the subject-object pair, although it has never seen the ground-truth location for the object entity during training.

\subsection{Localizing relationships in Visual Genome}\label{sec:exp_visgeno}

We also evaluate our method on the Visual Genome dataset \cite{krishna2016visual}, which contains relationship expressions annotated over pairs of objects, such as ``computer on top of table'' and ``person wearing shirt''.

On the relationship annotations in Visual Genome, given an image and an expression like ``man wearing hat'', we evaluate our method in two test scenarios: retrieving the \textit{subject} region (``man'') and retrieving the \textit{subject-object pair} (both ``man'' and ``hat''). In our experiment, we take the bounding boxes of all the annotated entities in each image (around 35 per image) as candidate region set $B$ at both training and test time, and extract visual features for each region from fc7 output of a Faster-RCNN VGG-16 network \cite{ren2015faster} pretrained on MSCOCO detection dataset \cite{lin2014microsoft}. The input images are first forwarded through the convolutional layers of the network, and the features of each image region are extracted by ROI-pooling over the convolutional feature map, followed by subsequent fully connected layers. We use the same training, validation and test split as in \cite{johnson2015densecap}.

Since there are ground-truth annotations for both subject region and object region in this dataset, we experiment with two training supervision settings: (1) \textit{weak supervision} by only providing the ground-truth region of the subject entity at training time (\textbf{subject-GT} in Table \ref{tab:results_visgeno}) and optimizing unary subject score $s_{subj}$ with Eqn. \ref{eqn:weak_supervision} and (2) \textit{strong supervision} by providing the ground-truth region pair of both subject and object entities at training time (\textbf{subject-object-GT} in Table \ref{tab:results_visgeno}) and optimizing pairwise score $s_{pair}$ with Eqn. \ref{eqn:strong_supervision}. 

Similar to the experiment on the synthetic dataset in Sec. \ref{sec:exp_shape}, we also train a baseline model that only looks at local appearance and spatial properties but ignores pairwise relationships. For the first evaluation scenario of retrieving the subject region, we train a baseline model using a localization module only by optimizing its output $s_{loc}$ for ground-truth subject region with softmax loss (the same training supervision as subject-GT). For the second scenario of retrieving the subject-object pair, we train two such baseline models optimized with subject ground-truth and object ground-truth respectively, to localize of the subject region and object region separately with each model and at test time combine the predicted subject region and predicted object region from each model be the subject-object pair (same training supervision as subject-object-GT).

\begin{table}[t]
\small
\begin{center}
\begin{tabular}{|l|c|c|c|}
\hline
Method & training supervision & P@1-subj & P@1-pair \\
\hline
baseline & subject-GT & 41.20\% & - \\
baseline & subject-object-GT & - & 23.37\% \\
our full model & subject-GT & 43.81\% & 26.56\% \\
our full model & subject-object-GT & \textbf{44.24\%} & \textbf{28.52\%} \\
\hline
\end{tabular}
\end{center}
\caption{Performance of our model on relationship expressions in Visual Genome dataset. See Sec. \ref{sec:exp_visgeno} for details.}
\label{tab:results_visgeno}
\end{table}

\begin{figure*}[t]
\begin{center}
\begin{tabularx}{\linewidth}{YYY|YYY}
ground-truth & our prediction & attention weights & ground-truth & our prediction & attention weights \\ \hline
\end{tabularx}
\begin{tabularx}{\linewidth}{Y|Y}
\small{expression=\textit{``tennis player wears shorts''}} & \small{expression=\textit{``building behind bus''}} \\
\includegraphics[trim = 25mm 19mm 20mm 15mm, clip=true,width=0.33\linewidth]{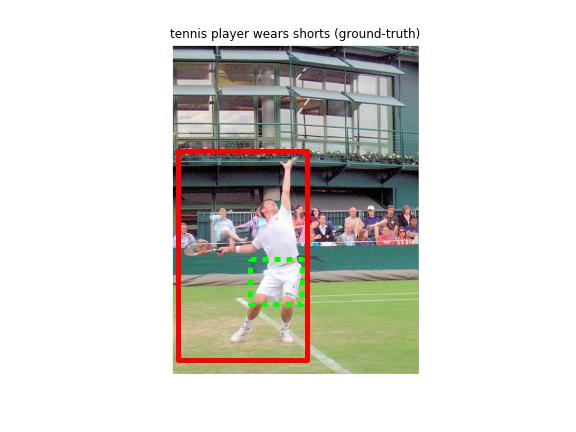}\includegraphics[trim = 25mm 19mm 20mm 15mm, clip=true,width=0.33\linewidth]{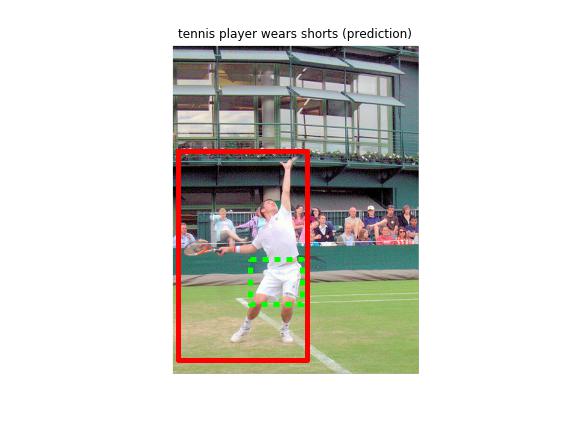}\includegraphics[trim = 0mm 25mm 105mm 55mm, clip=true,width=0.33\linewidth]{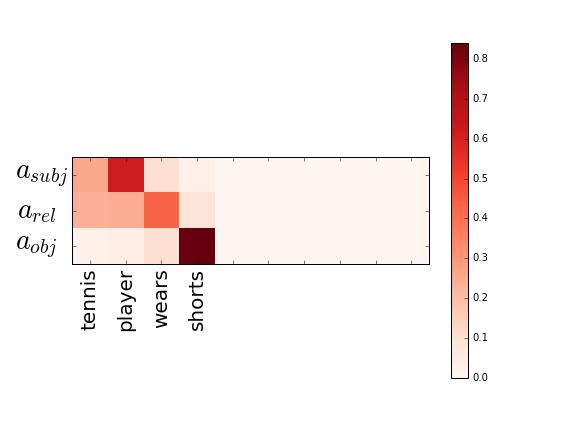} &
\includegraphics[trim = 25mm 19mm 20mm 15mm, clip=true,width=0.33\linewidth]{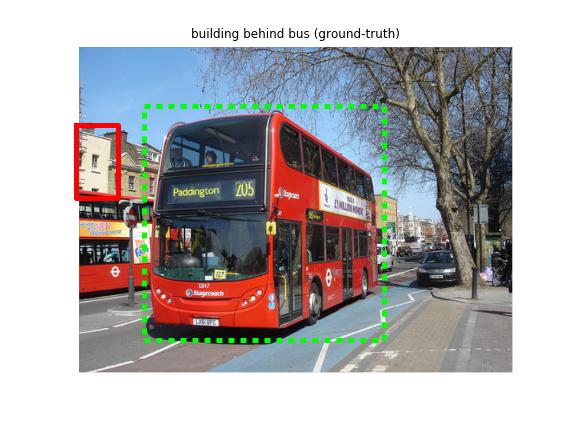}\includegraphics[trim = 25mm 19mm 20mm 15mm, clip=true,width=0.33\linewidth]{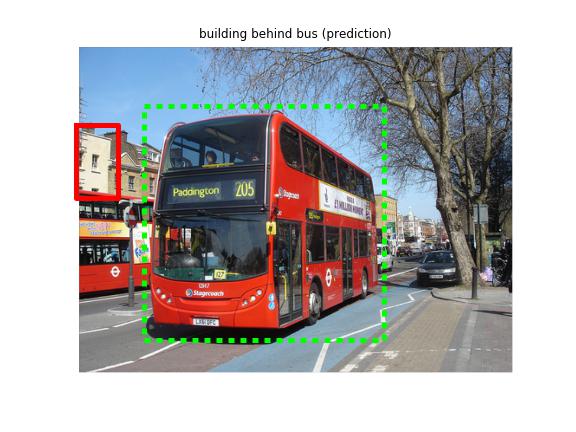}\includegraphics[trim = 0mm 25mm 105mm 55mm, clip=true,width=0.33\linewidth]{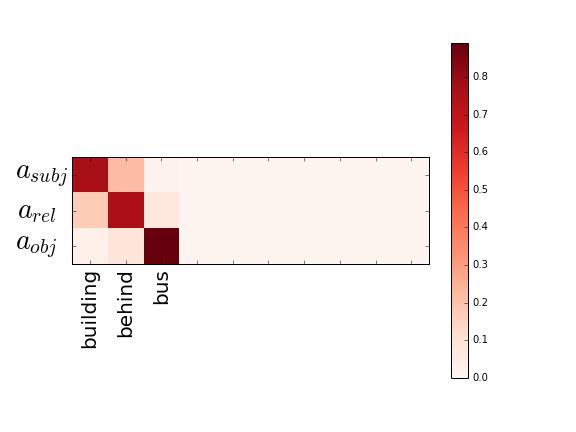} \\

\small{expression=\textit{``car has tail light''}} & \small{expression=\textit{``window on front of building''}} \\
\includegraphics[trim = 25mm 19mm 20mm 15mm, clip=true,width=0.33\linewidth]{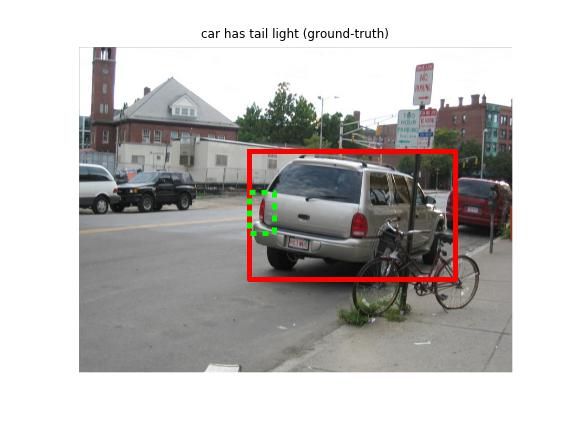}\includegraphics[trim = 25mm 19mm 20mm 15mm, clip=true,width=0.33\linewidth]{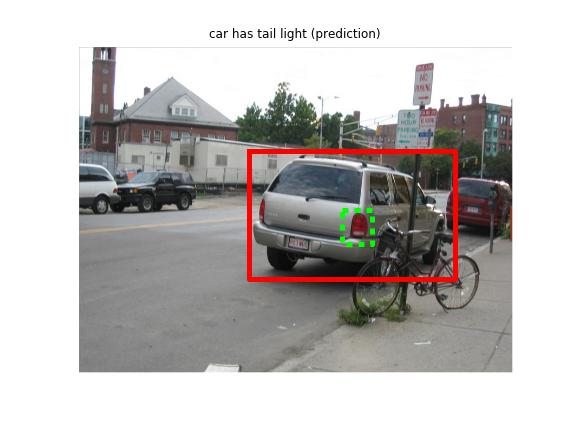}\includegraphics[trim = 0mm 25mm 105mm 55mm, clip=true,width=0.33\linewidth]{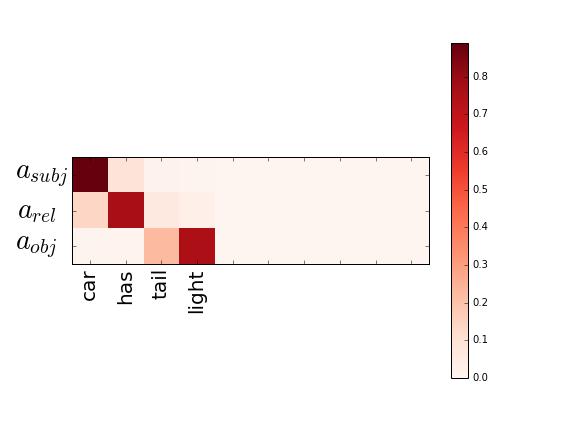} &
\includegraphics[trim = 25mm 19mm 20mm 15mm, clip=true,width=0.33\linewidth]{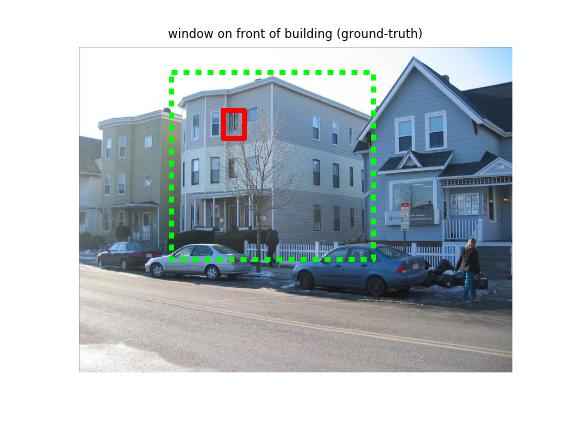}\includegraphics[trim = 25mm 19mm 20mm 15mm, clip=true,width=0.33\linewidth]{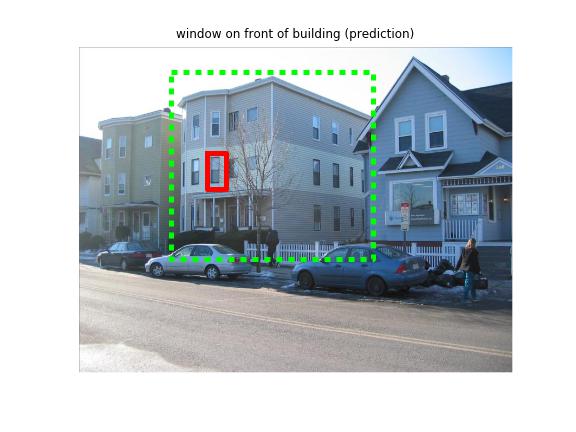}\includegraphics[trim = 0mm 25mm 105mm 55mm, clip=true,width=0.33\linewidth]{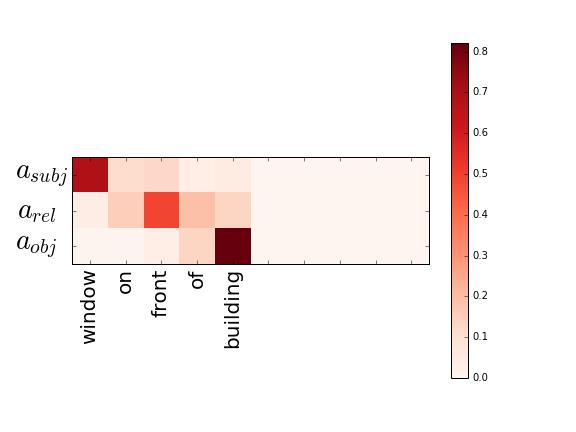} \\

\small{expression=\textit{``business name on sign''}} & \small{expression=\textit{``board on top of store''}} \\
\includegraphics[trim = 25mm 19mm 20mm 15mm, clip=true,width=0.33\linewidth]{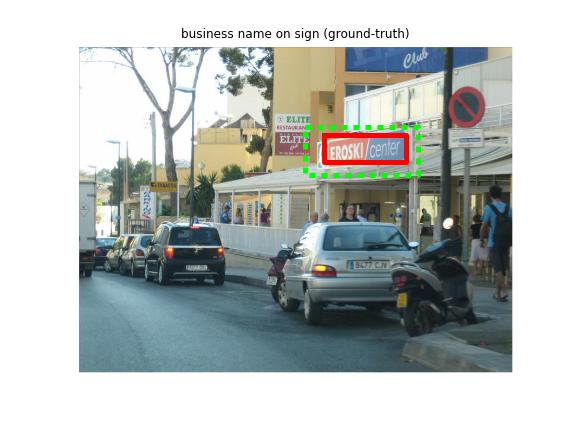}\includegraphics[trim = 25mm 19mm 20mm 15mm, clip=true,width=0.33\linewidth]{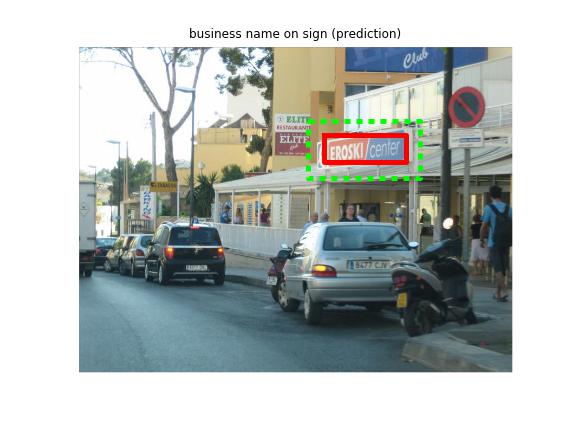}\includegraphics[trim = 0mm 25mm 105mm 55mm, clip=true,width=0.33\linewidth]{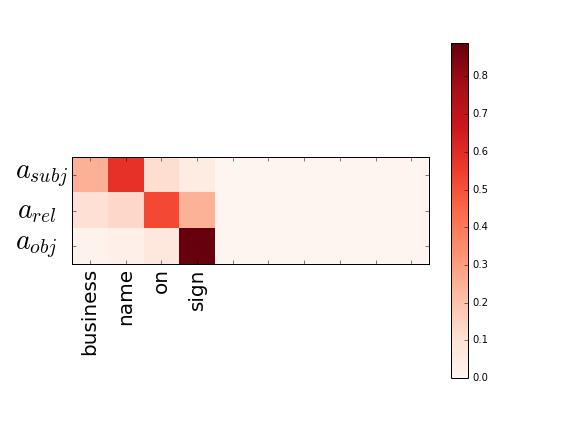} &
\includegraphics[trim = 25mm 19mm 20mm 15mm, clip=true,width=0.33\linewidth]{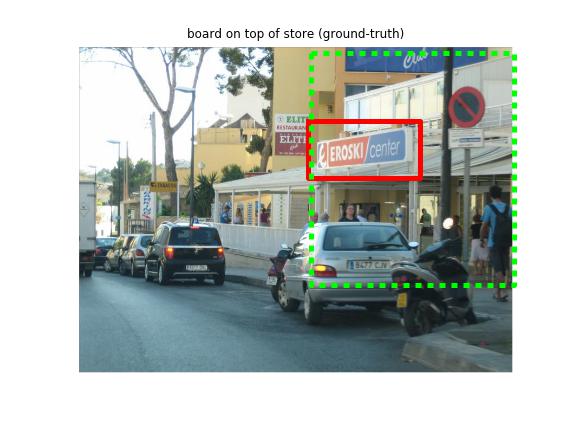}\includegraphics[trim = 25mm 19mm 20mm 15mm, clip=true,width=0.33\linewidth]{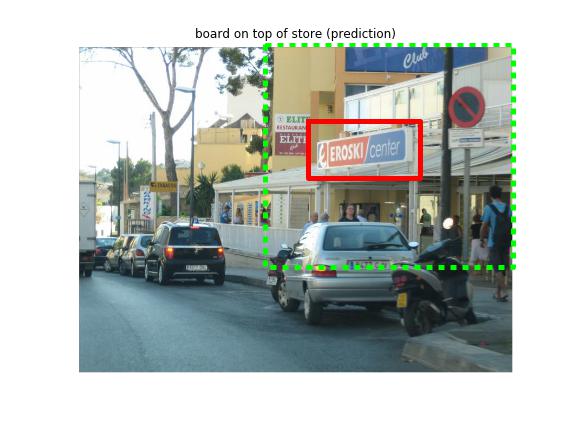}\includegraphics[trim = 0mm 25mm 105mm 55mm, clip=true,width=0.33\linewidth]{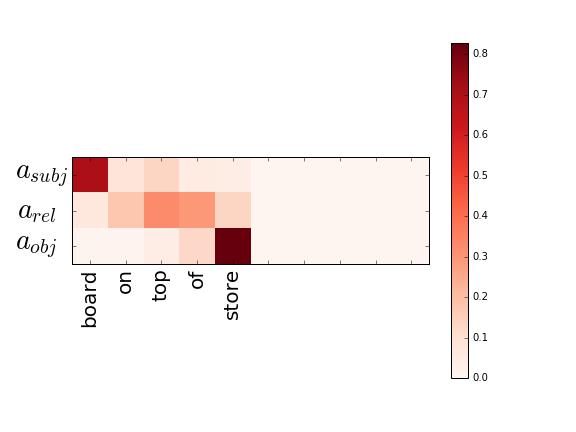} \\

\small{expression=\textit{``wine bottle next to glasses''}} & \small{expression=\textit{``chairs around table''}} \\
\includegraphics[trim = 25mm 19mm 20mm 15mm, clip=true,width=0.33\linewidth]{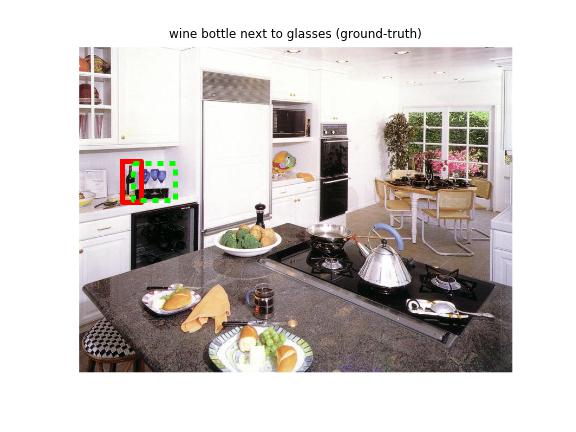}\includegraphics[trim = 25mm 19mm 20mm 15mm, clip=true,width=0.33\linewidth]{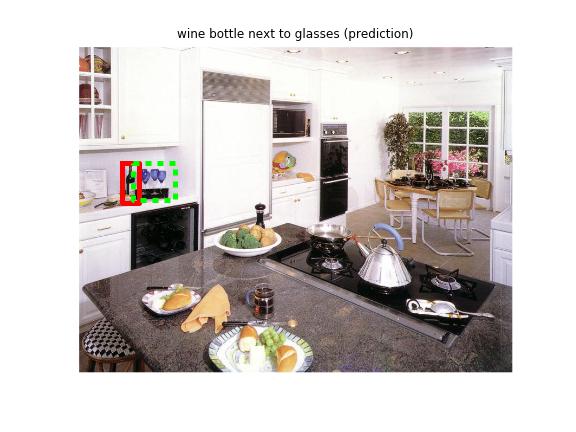}\includegraphics[trim = 0mm 25mm 105mm 55mm, clip=true,width=0.33\linewidth]{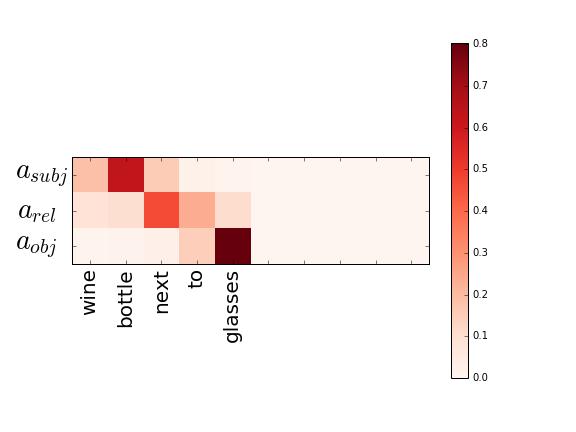} &
\includegraphics[trim = 25mm 19mm 20mm 15mm, clip=true,width=0.33\linewidth]{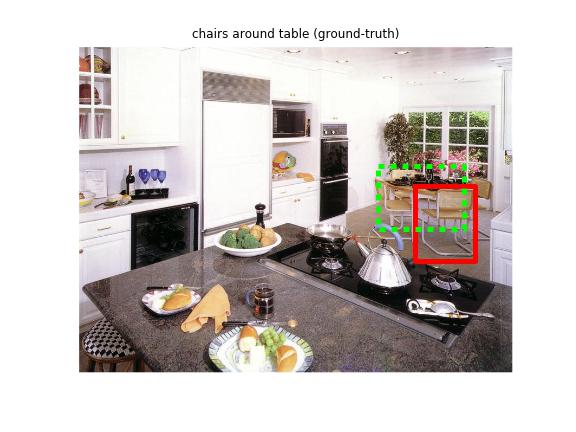}\includegraphics[trim = 25mm 19mm 20mm 15mm, clip=true,width=0.33\linewidth]{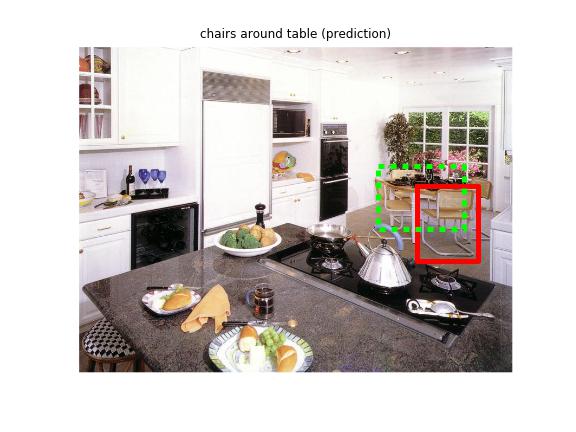}\includegraphics[trim = 0mm 25mm 105mm 55mm, clip=true,width=0.33\linewidth]{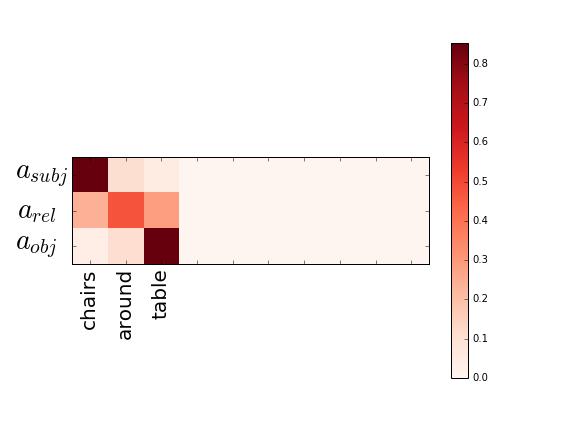} \\

\small{expression=\textit{``marker on top of ledge''}} & \small{expression=\textit{``chair next to table''}} \\
\includegraphics[trim = 25mm 19mm 20mm 15mm, clip=true,width=0.33\linewidth]{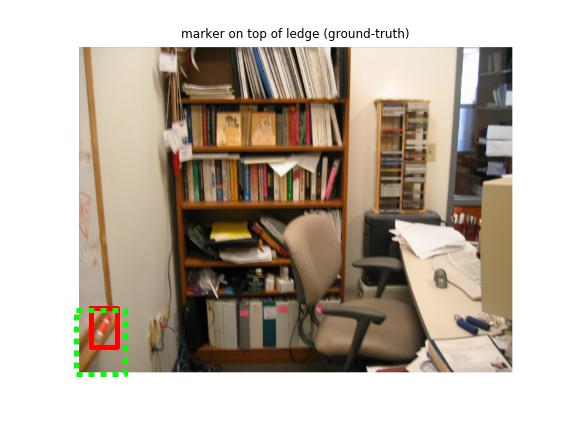}\includegraphics[trim = 25mm 19mm 20mm 15mm, clip=true,width=0.33\linewidth]{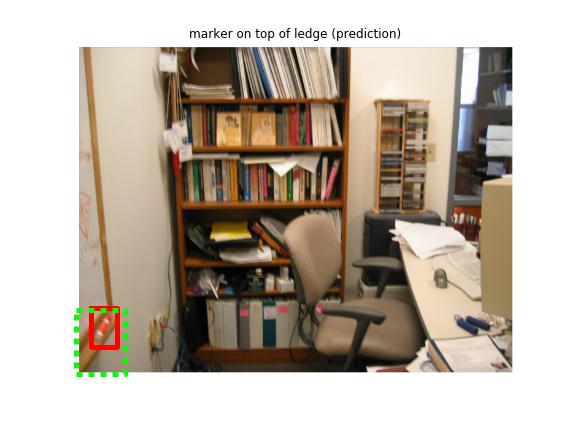}\includegraphics[trim = 0mm 25mm 105mm 55mm, clip=true,width=0.33\linewidth]{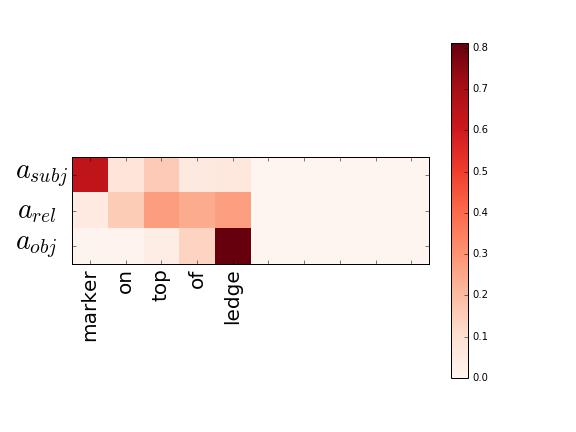} &
\includegraphics[trim = 25mm 19mm 20mm 15mm, clip=true,width=0.33\linewidth]{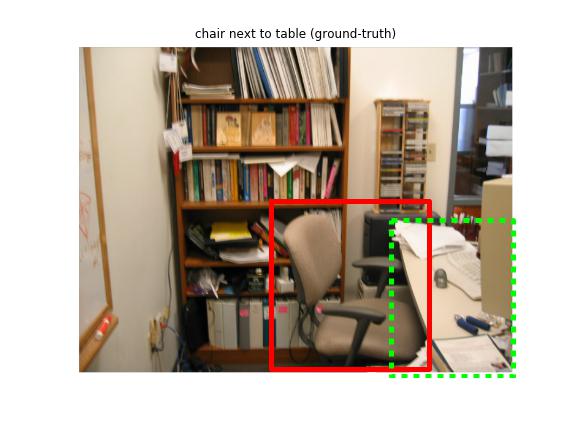}\includegraphics[trim = 25mm 19mm 20mm 15mm, clip=true,width=0.33\linewidth]{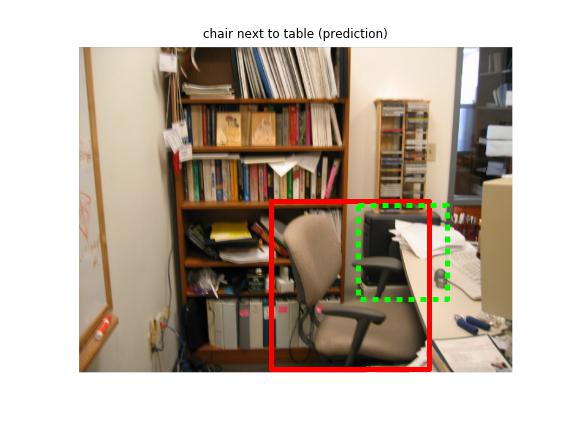}\includegraphics[trim = 0mm 25mm 105mm 55mm, clip=true,width=0.33\linewidth]{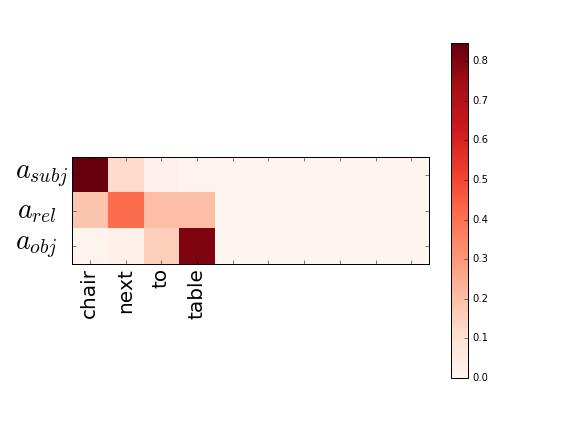}
\end{tabularx}
\end{center}
\caption{Visualization of grounded relationship expressions in the Visual Genome dataset, trained with weak supervision (subject-GT). In each example, the first and the second column show ground-truth region pairs and our predicted region pairs respectively (subject in \textcolor{red}{red} solid box and object in \textcolor{green}{green} dashed box). The third column visualizes attention weights in Eqn. \ref{eqn:attention_1}--\ref{eqn:attention_3} for subject, relationship and object (darker is higher).}
\label{fig:vis_visgeno}
\end{figure*}

We evaluate with top-1 precision (P@1), which is the percentage of test instances where the top scoring prediction matches the ground-truth in each image (P@1-subj for predicted subject regions matching subject ground-truth in the first scenario, and P@1-pair for predicted subject and object regions both matching the ground-truth in the second scenario). The results are summarized in Table \ref{tab:results_visgeno}, where it can be seen that our full model outperforms the baseline using only localization modules in both evaluation scenarios. Note that in the second evaluation scenario of retrieving subject-object pairs, our weakly supervised model still outperforms the baseline trained with strong supervision.

Figure \ref{fig:vis_visgeno} shows some examples of our model trained with weak supervision (subject-GT) and attention weights in Eqn. \ref{eqn:attention_1}--\ref{eqn:attention_3}. It can be seen that even with weak supervision, our model still generates reasonable attention weights over words for subject, relationship and object.

\subsection{Grounding referential expressions in images}\label{sec:exp_refgoog}

We apply our model to the Google-Ref dataset \cite{mao2016generation}, a benchmark dataset for grounding referential expressions. As this dataset does not explicitly contain subject-object pair annotation for the referential expressions, we train our model with weak supervision (Eqn. \ref{eqn:weak_supervision}) by optimizing the subject score $s_{subj}$ using the expression-level region ground-truth. The candidate bounding box set $B$ at both training and test time are all the annotated entities in the image (which is the ``Ground-Truth'' evaluation setting in \cite{mao2016generation}). As in Sec. \ref{sec:exp_visgeno}, fc7 output of a MSCOCO-pretrained Faster-RCNN VGG-16 network is used for visual feature extraction. Similar to Sec. \ref{sec:exp_shape}, we also train a GroundeR-like \cite{rohrbach2016grounding} baseline model with localization module which looks only at a region's local features.

\begin{table}[t]
\begin{center}
\begin{tabular}{|l|c|}
\hline
Method & P@1 \\
\hline
Mao \etal \cite{mao2016generation} & 60.7\% \\
Yu \etal \cite{yu2016modeling} & 64.0\% \\
Nagaraja \etal \cite{nagaraja2016modeling} & 68.4\% \\
\hline
baseline (loc module) & 66.5\% \\
our model (w/ external parser) & 53.5\% \\
our full model & \textbf{69.3\%} \\
\hline
\end{tabular}
\end{center}
\caption{Top-1 precision of our model and previous methods on Google-Ref dataset. See Sec. \ref{sec:exp_refgoog} for details.}
\label{tab:results_refgoog}
\end{table}

In addition, instead of learning a linguistic analysis end-to-end as in Sec.\ \ref{sec:lang_representation}, we also experiment with parsing the expression using the Stanford Parser \cite{zhu2013fast,stanfordparser}. An expression is parsed into subject, relationship and object component according to the constituency tree, and the components are encoded into vectors $q_{subj}$, $q_{rel}$ and $q_{obj}$ using three separate LSTM encoders, similar to the baseline and \cite{rohrbach2016grounding}.

Following \cite{mao2016generation}, we evaluate on this dataset using the top-1 precision (P@1) metric, which is the fraction of the highest scoring subject region matching the ground-truth for the expression. Table \ref{tab:results_refgoog} shows the performance of our model, baseline model and previous work. Note that all the methods are trained with the same weak supervision (only a ground-truth subject region). It can be seen that by incorporating inter-object relationships, our full model outperforms the baseline using only localization modules, and works better than previous state-of-the-art methods.

Additionally, replacing the learned expression parsing and language representation in Sec.\ \ref{sec:lang_representation} with an external parser (``our model w/ external parser'' in Table \ref{tab:results_refgoog}) leads to a significant performance drop. We find that this is mainly because existing parsers are not specifically tuned for the referring expression task---as noted in Sec.\ \ref{sec:lang_representation}, expressions like \emph{chair on the left of the table} are parsed as (\emph{chair}, \emph{on}, \emph{the left of the table}) rather than the desired triplet (\emph{chair}, \emph{on the left of}, \emph{the table}). In our full model, the language representation is end-to-end optimized with other parts, while it is hard to jointly optimize an external language parser like \cite{zhu2013fast} for this task.

Figure \ref{fig:vis_refgoog} shows some example results on this dataset. It can be seen that although weakly supervised, our model not only grounds the subject region correctly (solid box), but also finds reasonable regions (dashed box) for the object entity.

\subsection{Answering pointing questions in Visual-7W}\label{sec:exp_visual7w}

\begin{table}[t]
\begin{center}
\begin{tabular}{|l|c|}
\hline
Method & Accuracy \\
\hline
Zhu \etal \cite{zhu2015visual7w} & 56.10\% \\
\hline
baseline (loc module) & 71.61\% \\
our model (w/ external parser) & 61.66\% \\
our full model & \textbf{72.53\%} \\
\hline
\end{tabular}
\end{center}
\caption{Accuracy of our model and previous methods on the pointing questions in Visual-7W dataset. See Sec. \ref{sec:exp_visual7w} for details.}
\label{tab:results_visual7w}
\end{table}

Finally, we evaluate our method on the multiple choice pointing questions (\ie ``which'' questions) in visual question answering on the Visual-7W dataset \cite{zhu2015visual7w}. Given an image and a question like ``which tomato slice is under the knife'', the task is to select the corresponding region from a few choice regions (4 choices in this dataset) as answer. Since this task is closely related to grounding referential expressions, our model can be trained in the same way as in Sec. \ref{sec:exp_refgoog} to score each choice region using subject score $s_{subj}$ and pick the highest scoring choice as answer.

As before, we train our model with weak supervision through Eqn. \ref{eqn:weak_supervision} and use a MSCOCO-pretrained Faster-RCNN VGG-16 network for visual feature extraction. Here we use two different candidate bounding box sets $B_{subj}$ and $B_{obj}$ of the subject regions (the choices) and the object regions, where $B_{subj}$ is the 4 choice bounding boxes, and $B_{obj}$ is the set of 300 proposal bounding boxes extracted using RPN in Faster-RCNN \cite{ren2015faster}. Similar to Sec. \ref{sec:exp_refgoog}, we also train a baseline model using only a localization module to score each choice based only on its local appearance and spatial properties, and a truncated model that uses the Stanford parser \cite{zhu2013fast,stanfordparser} for expression parsing and language representation.

\begin{figure*}
\center
\begin{tabularx}{\linewidth}{YY|YY|YY}
ground-truth & our prediction & ground-truth & our prediction & ground-truth & our prediction \\ \hline
\end{tabularx}
\begin{tabularx}{\linewidth}{Y|Y|Y}
\footnotesize{expression=\textit{``a bear lying to the right of another bear''}} &
\footnotesize{expression=\textit{``man in sunglasses walking towards two talking men''}} &
\footnotesize{expression=\textit{``a picnic table that has a bottle of water sitting on it''}} \\
\includegraphics[trim = 25mm 19mm 20mm 15mm, clip=true,width=0.5\linewidth]{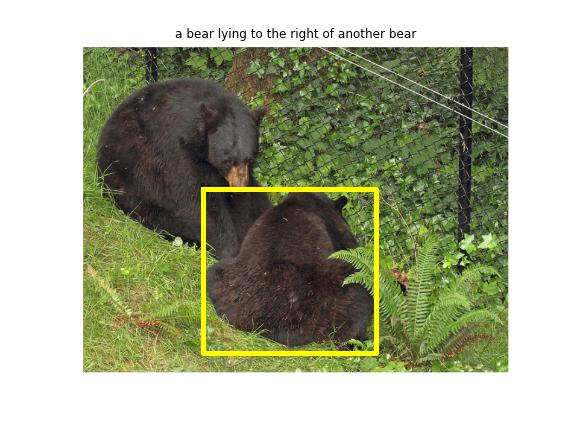}\includegraphics[trim = 25mm 19mm 20mm 15mm, clip=true,width=0.5\linewidth]{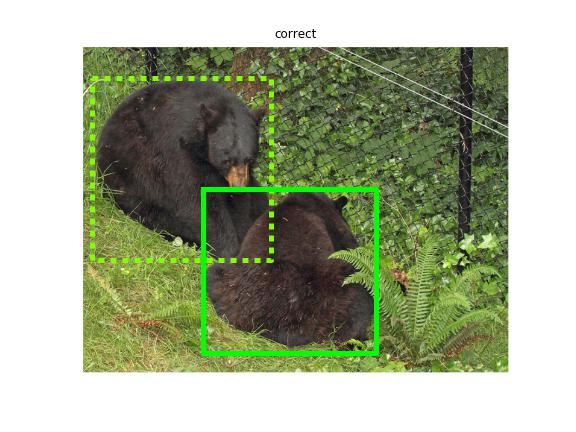} &
\includegraphics[trim = 25mm 19mm 20mm 21mm, clip=true,width=0.5\linewidth]{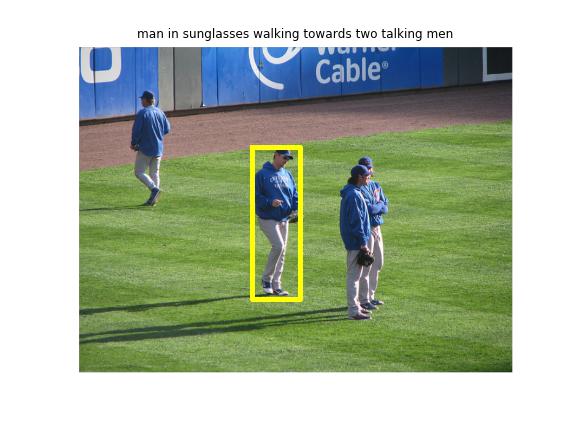}\includegraphics[trim = 25mm 19mm 20mm 21mm, clip=true,width=0.5\linewidth]{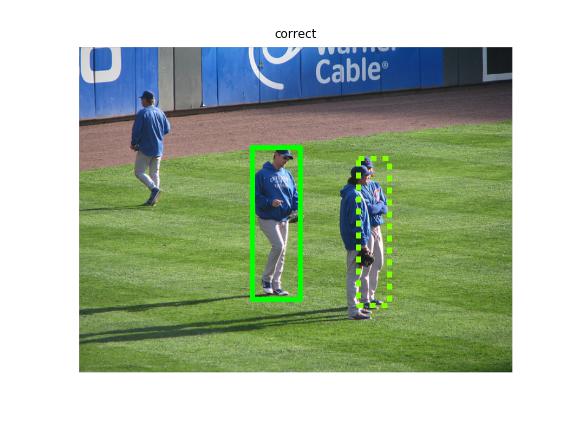} &
\includegraphics[trim = 25mm 19mm 20mm 21mm, clip=true,width=0.5\linewidth]{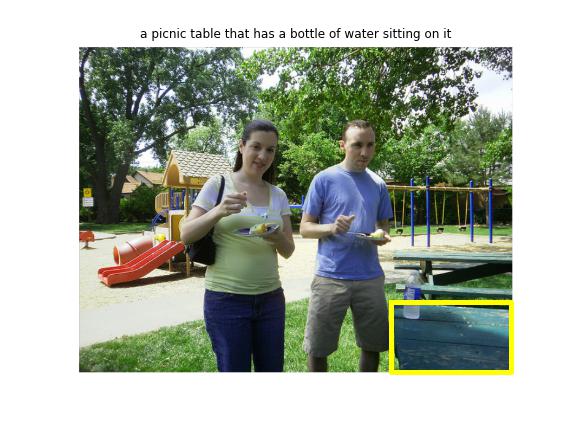}\includegraphics[trim = 25mm 19mm 20mm 21mm, clip=true,width=0.5\linewidth]{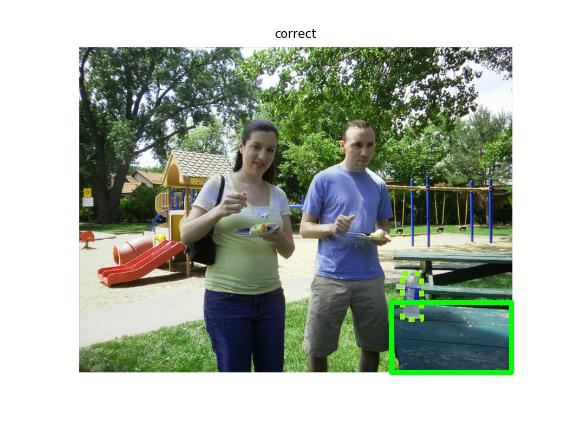} \\
\footnotesize{correct} &
\footnotesize{correct} &
\footnotesize{correct} \\
\hline

\footnotesize{expression=\textit{``woman in a cream colored wedding dress cutting cake''}} &
\footnotesize{expression=\textit{``a man going before a lady carrying a cellphone''}} &
\footnotesize{expression=\textit{``pizza slice not eaten''}} \\
\includegraphics[trim = 25mm 19mm 20mm 15mm, clip=true,width=0.5\linewidth]{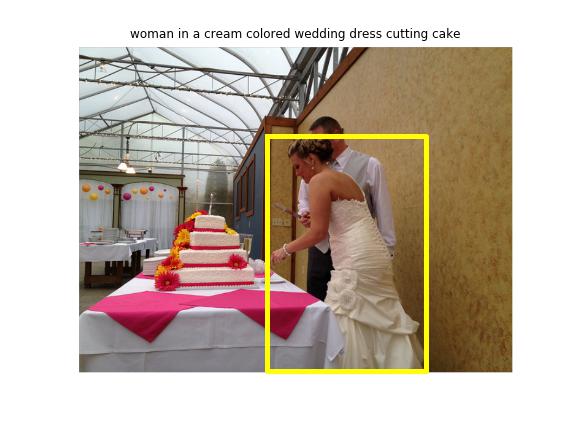}\includegraphics[trim = 25mm 19mm 20mm 15mm, clip=true,width=0.5\linewidth]{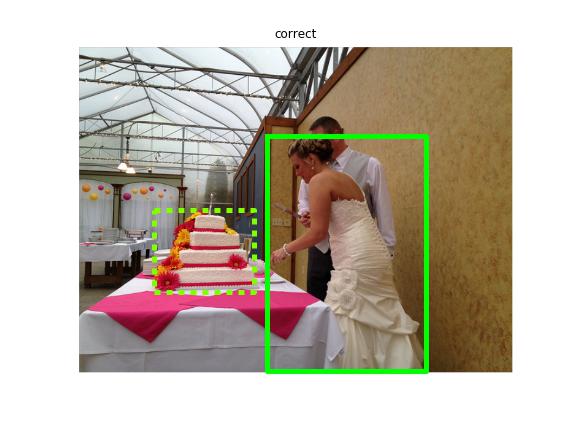} &
\includegraphics[trim = 25mm 19mm 20mm 15mm, clip=true,width=0.5\linewidth]{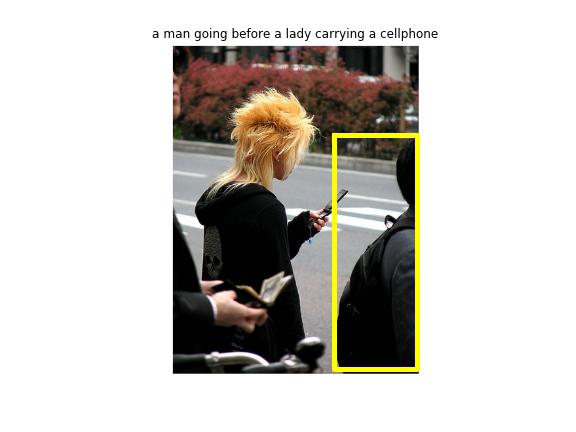}\includegraphics[trim = 25mm 19mm 20mm 15mm, clip=true,width=0.5\linewidth]{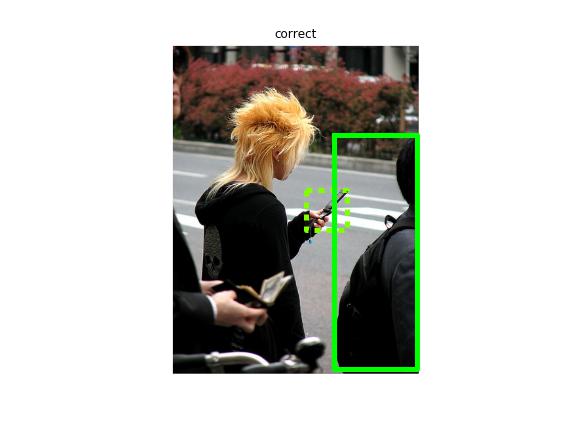} &
\includegraphics[trim = 25mm 19mm 20mm 21mm, clip=true,width=0.5\linewidth]{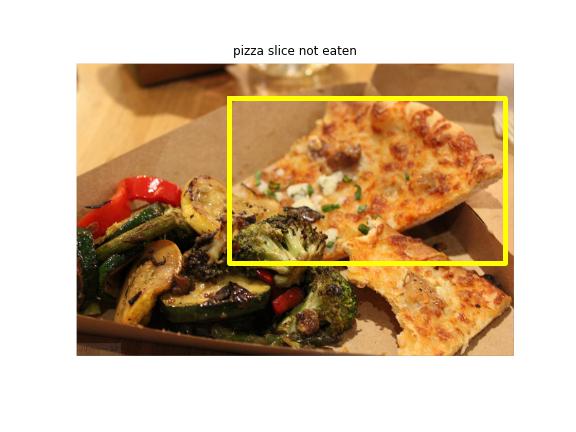}\includegraphics[trim = 25mm 19mm 20mm 21mm, clip=true,width=0.5\linewidth]{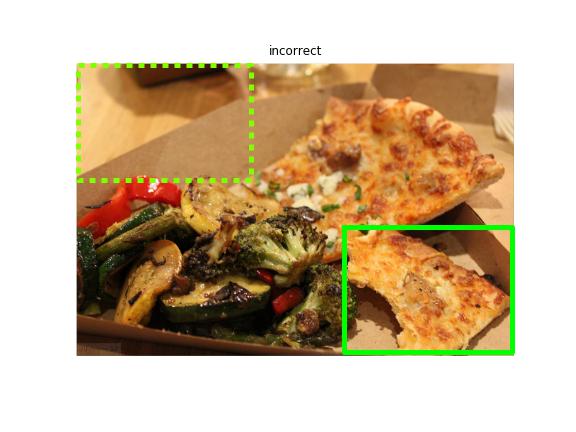} \\
\footnotesize{correct} &
\footnotesize{correct} &
\footnotesize{incorrect} \\
\hline

\footnotesize{expression=\textit{``a full grown brown bear near a young bear''}} &
\footnotesize{expression=\textit{``black dog standing on all four legs''}} &
\footnotesize{expression=\textit{``chair being sat in by a man''}} \\
\includegraphics[trim = 25mm 19mm 20mm 28mm, clip=true,width=0.5\linewidth]{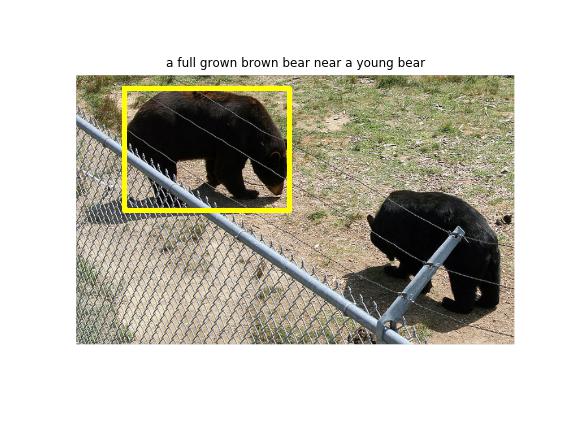}\includegraphics[trim = 25mm 19mm 20mm 28mm, clip=true,width=0.5\linewidth]{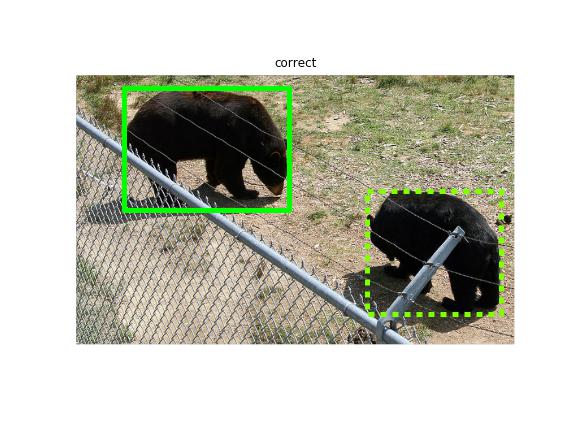} &
\includegraphics[trim = 25mm 19mm 20mm 21mm, clip=true,width=0.5\linewidth]{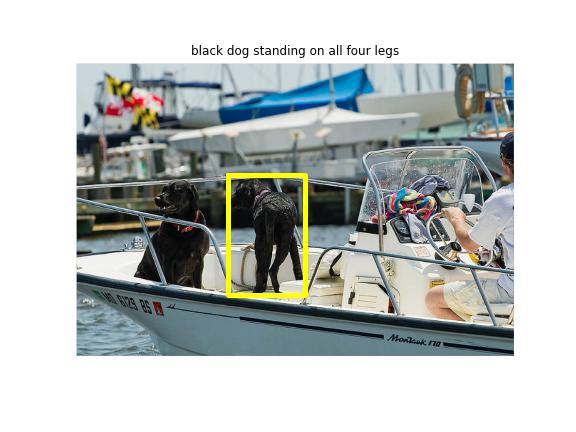}\includegraphics[trim = 25mm 19mm 20mm 21mm, clip=true,width=0.5\linewidth]{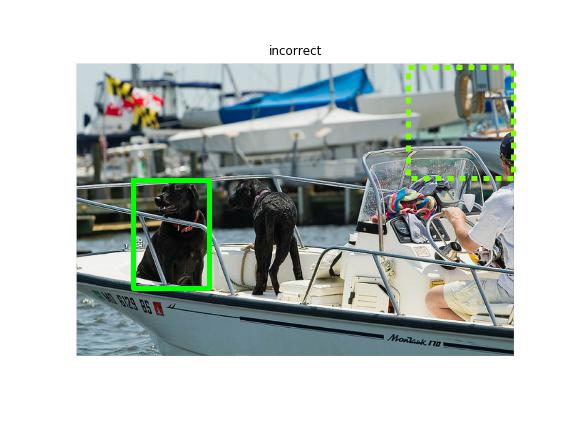} &
\includegraphics[trim = 25mm 19mm 20mm 21mm, clip=true,width=0.5\linewidth]{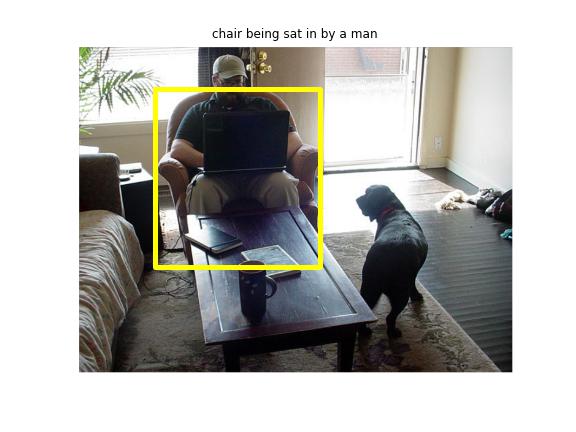}\includegraphics[trim = 25mm 19mm 20mm 21mm, clip=true,width=0.5\linewidth]{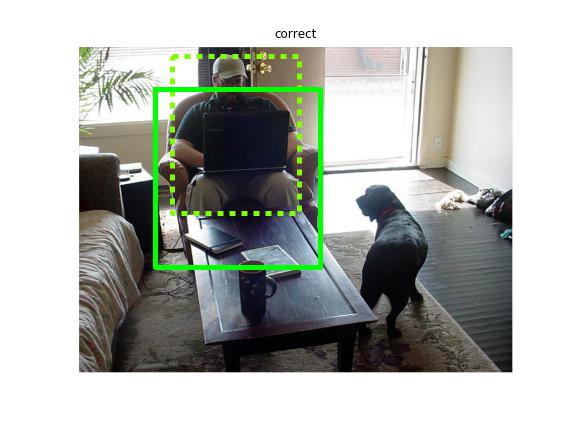} \\
\footnotesize{correct} &
\footnotesize{incorrect} &
\footnotesize{correct} \\
\end{tabularx}
\caption{Examples of referential expressions in the Google-Ref dataset. The left column shows the ground-truth region and the right column shows the grounded subject region (our prediction) in solid box and the grounded object region in dashed box. A prediction is labeled as correct if the predicted subject region matches the ground-truth region.}
\label{fig:vis_refgoog}

\vspace{0.1 cm}

\begin{tabularx}{\linewidth}{YY|YY|YY}
ground-truth & our prediction & ground-truth & our prediction & ground-truth & our prediction \\ \hline
\end{tabularx}
\begin{tabularx}{\linewidth}{Y|Y|Y}
\footnotesize{question=\textit{``Which wine glass is in the man's hand?''}} &
\footnotesize{question=\textit{``Which person is wearing a helmet?''}} &
\footnotesize{question=\textit{``Which mouse is on a pad by computer?''}} \\
\includegraphics[trim = 25mm 19mm 20mm 28mm, clip=true,width=0.5\linewidth]{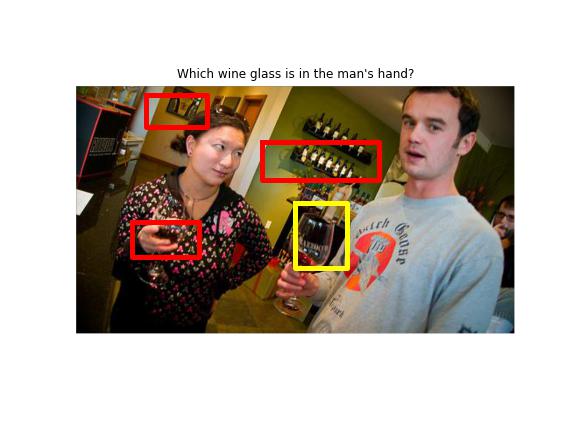}\includegraphics[trim = 25mm 19mm 20mm 28mm, clip=true,width=0.5\linewidth]{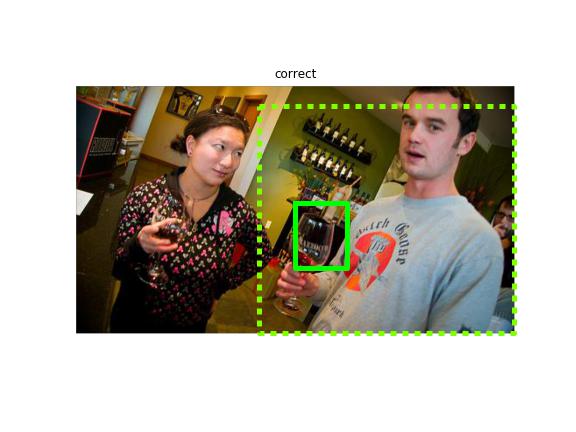} &
\includegraphics[trim = 25mm 19mm 20mm 21mm, clip=true,width=0.5\linewidth]{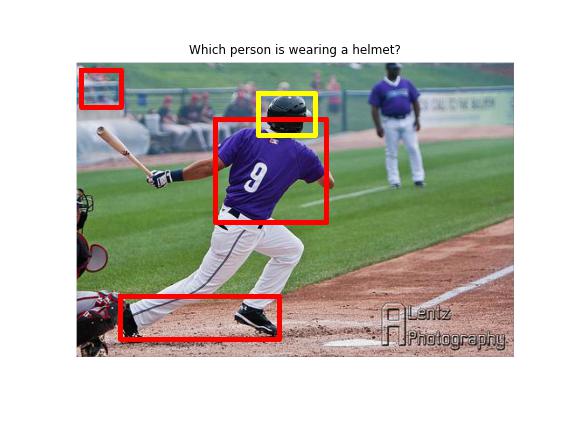}\includegraphics[trim = 25mm 19mm 20mm 21mm, clip=true,width=0.5\linewidth]{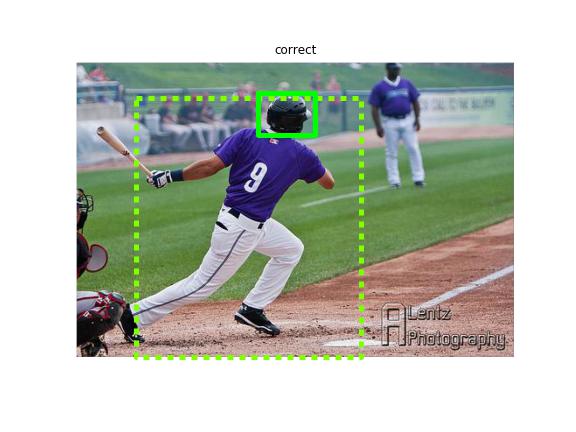} &
\includegraphics[trim = 25mm 19mm 20mm 15mm, clip=true,width=0.5\linewidth]{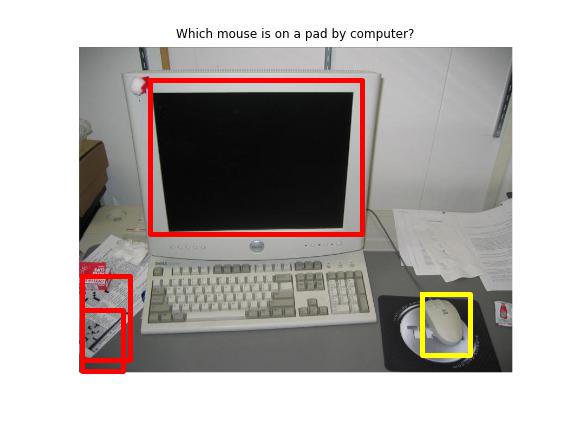}\includegraphics[trim = 25mm 19mm 20mm 15mm, clip=true,width=0.5\linewidth]{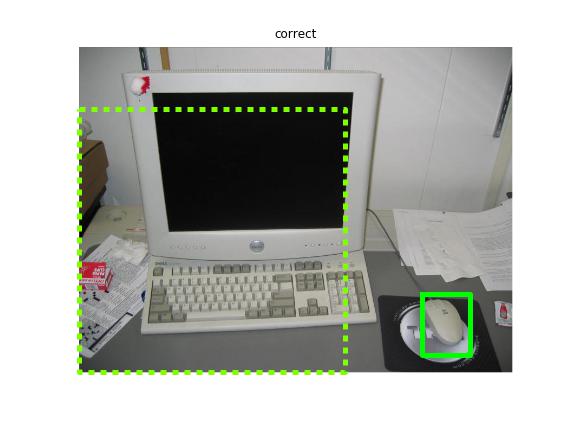} \\
\footnotesize{correct} &
\footnotesize{correct} &
\footnotesize{correct} \\
\hline

\footnotesize{question=\textit{``Which head is that of an adult giraffe?''}} &
\footnotesize{question=\textit{``Which pants belong to the man closest to the train?''}} &
\footnotesize{question=\textit{``Which white pillow is leftmost on the bed?''}} \\
\includegraphics[trim = 25mm 19mm 20mm 21mm, clip=true,width=0.5\linewidth]{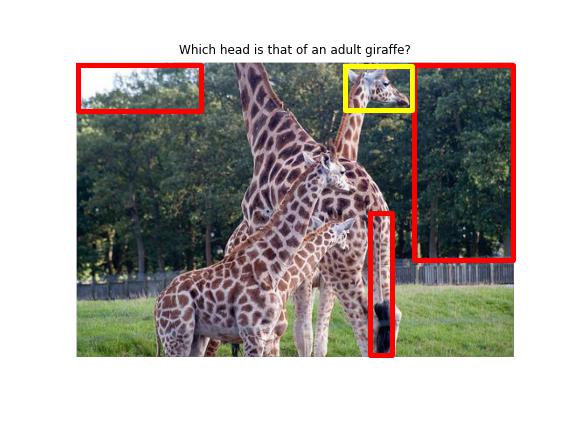}\includegraphics[trim = 25mm 19mm 20mm 21mm, clip=true,width=0.5\linewidth]{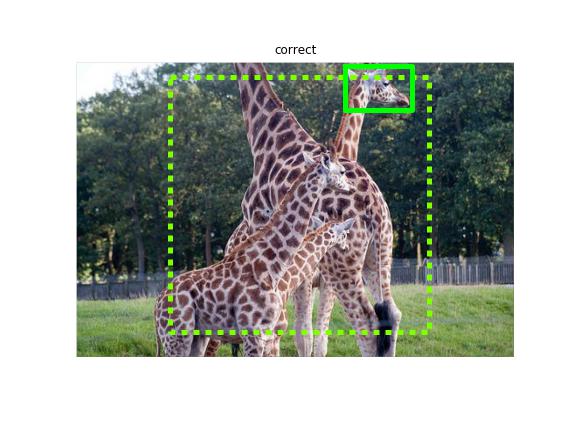} &
\includegraphics[trim = 25mm 19mm 20mm 21mm, clip=true,width=0.5\linewidth]{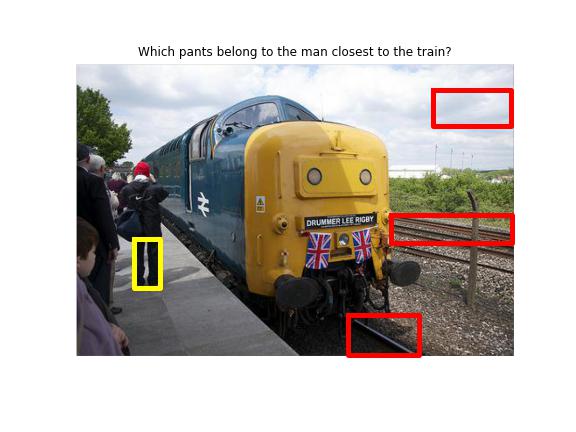}\includegraphics[trim = 25mm 19mm 20mm 21mm, clip=true,width=0.5\linewidth]{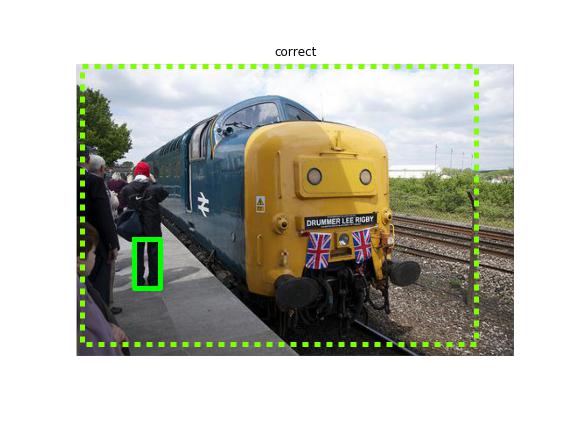} &
\includegraphics[trim = 25mm 19mm 20mm 15mm, clip=true,width=0.5\linewidth]{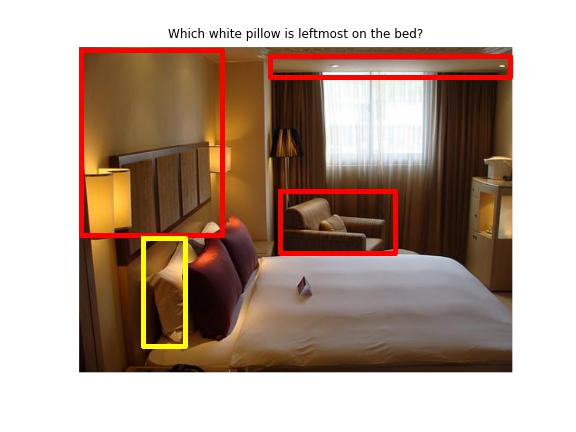}\includegraphics[trim = 25mm 19mm 20mm 15mm, clip=true,width=0.5\linewidth]{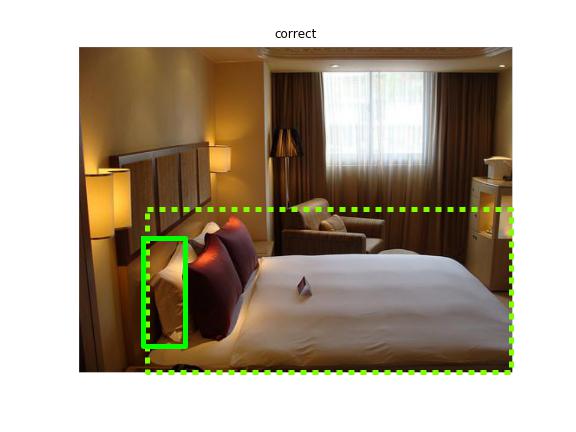} \\
\footnotesize{correct} &
\footnotesize{correct} &
\footnotesize{correct} \\
\hline

\footnotesize{question=\textit{``Which red shape is on a large white sign?''}} &
\footnotesize{question=\textit{``Which is not a pair of a living canine?''}} &
\footnotesize{question=\textit{``Which hand can be seen from under the umbrella?''}} \\
\includegraphics[trim = 25mm 19mm 20mm 15mm, clip=true,width=0.5\linewidth]{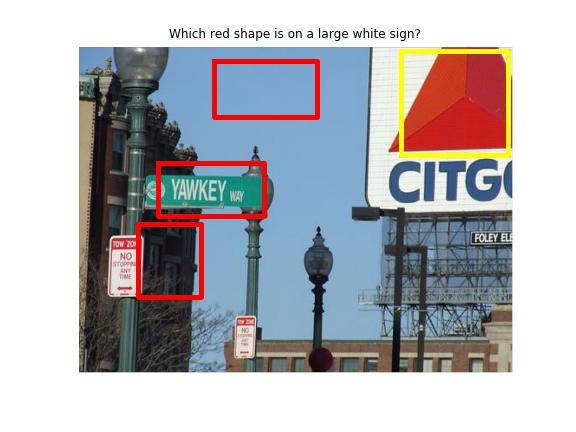}\includegraphics[trim = 25mm 19mm 20mm 15mm, clip=true,width=0.5\linewidth]{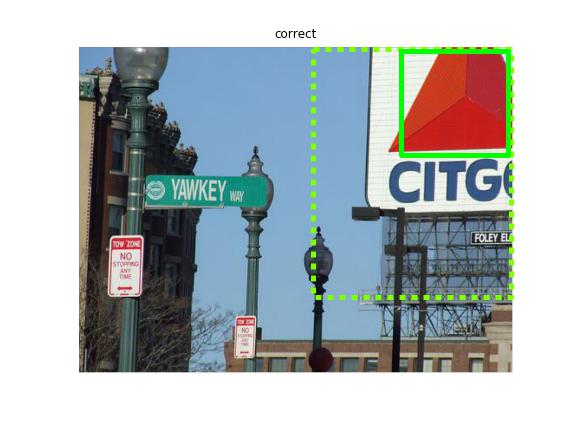} &
\includegraphics[trim = 25mm 19mm 20mm 15mm, clip=true,width=0.5\linewidth]{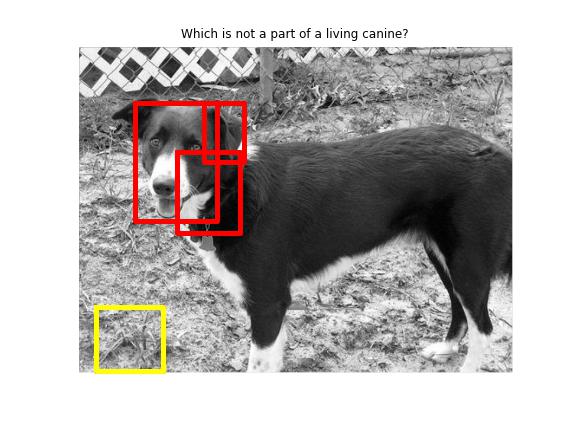}\includegraphics[trim = 25mm 19mm 20mm 15mm, clip=true,width=0.5\linewidth]{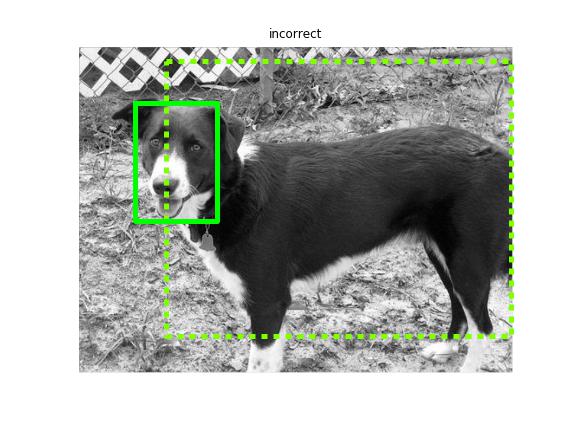} &
\includegraphics[trim = 25mm 19mm 20mm 15mm, clip=true,width=0.5\linewidth]{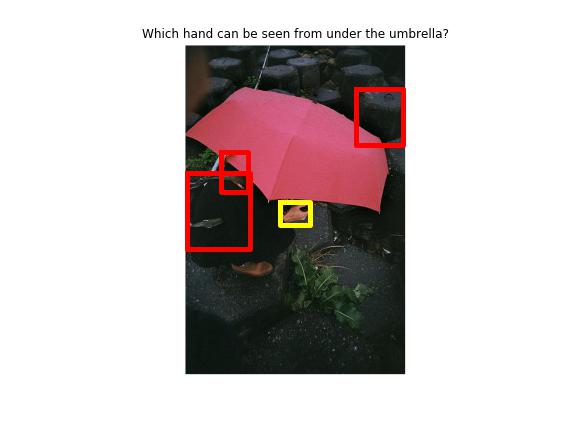}\includegraphics[trim = 25mm 19mm 20mm 15mm, clip=true,width=0.5\linewidth]{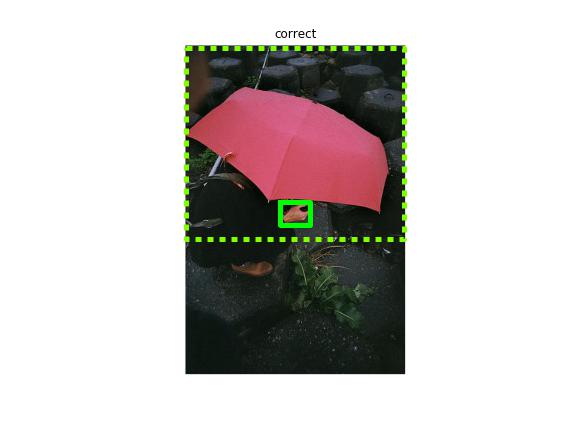} \\
\footnotesize{correct} &
\footnotesize{incorrect} &
\footnotesize{correct} \\

\end{tabularx}
\caption{Example pointing questions in the Visual-7W dataset. The left column shows the 4 multiple choices (ground-truth answer in \textcolor{bananayellow}{yellow}) and the right column shows the grounded subject region (predicted answer) in solid box and the grounded object region in dashed box. A prediction is labeled as correct if the predicted subject region matches the ground-truth region.}
\label{fig:vis_visual7w}
\end{figure*}

The results are shown in Table \ref{tab:results_visual7w}. It can be seen that our full model outperforms the baseline and the truncated model with an external parser, and achieves much higher accuracy than previous work \cite{zhu2015visual7w}. Figure \ref{fig:vis_visual7w} shows some question answering examples on this dataset.

\section{Conclusion}\label{sec:conclusion}

We have proposed Compositional Modular Networks, a novel end-to-end trainable model for handling relationships in referential expressions. Our model learns to parse input expressions with soft attention, and incorporates two types of modules that consider a region's local features and pairwise interaction between regions respectively. The model induces intuitive linguistic and visual analyses of referential expressions from only weak supervision, and experimental results demonstrate that our approach outperforms both natural baselines and state-of-the-art methods on multiple datasets.

\section*{Acknowledgements}
This work was supported by DARPA, AFRL, DoD MURI award N000141110688, NSF awards IIS-1427425, IIS-1212798 and IIS-1212928, NGA and the Berkeley Artificial Intelligence Research (BAIR) Lab. Jacob Andreas is supported by a Facebook graduate fellowship and a Huawei / Berkeley AI fellowship.

{\small
\bibliographystyle{ieee}
\bibliography{references}
}

\end{document}